\pdfoutput=1
\documentclass[nohyperref]{article}

\usepackage{microtype}
\usepackage{graphicx}
\usepackage{subfigure}
\usepackage{booktabs} 

\usepackage{hyperref}

\usepackage[accepted]{icml2023}

\usepackage{amsmath}
\usepackage{amssymb}
\usepackage{mathtools}
\usepackage{amsthm}
\usepackage{accessibility}
\usepackage{bbm}

\usepackage[capitalize,noabbrev]{cleveref}

\theoremstyle{plain}

\theoremstyle{definition}

\theoremstyle{remark}

\newcommand{\new}[1]{{\color{black} #1}}

\DeclareMathOperator{\sign}{sgn}

\usepackage[textsize=tiny]{todonotes}

\icmltitlerunning{Men Also Do Laundry: Multi-Attribute Bias Amplification}

\begin{document}

\twocolumn[
\icmltitle{Men Also Do Laundry: Multi-Attribute Bias Amplification}



\icmlsetsymbol{equal}{*}

\begin{icmlauthorlist}
\icmlauthor{Dora Zhao}{yyy}
\icmlauthor{Jerone T.~A. Andrews}{comp}
\icmlauthor{Alice Xiang}{yyy}
\end{icmlauthorlist}

\icmlaffiliation{yyy}{Sony AI, New York}
\icmlaffiliation{comp}{Sony AI, Tokyo}

\icmlcorrespondingauthor{Dora Zhao}{dora.zhao@sony.com}

\icmlkeywords{Machine Learning, ICML}

\vskip 0.3in
]

\newcommand{\smallsec}[1]{\vspace{0.15cm} \noindent {\bf #1.}}



\printAffiliationsAndNotice{}  

\begin{abstract}
    The phenomenon of \emph{bias amplification} occurs when models amplify training set biases at test time. Existing metrics measure bias amplification with respect to single annotated attributes (e.g., \texttt{computer}). However, large-scale datasets typically consist of instances with multiple attribute annotations (e.g., \{\texttt{computer}, \texttt{keyboard}\}). We demonstrate models can learn to exploit correlations with respect to multiple attributes, which are not accounted for by current metrics. Moreover, we show that current metrics can give the erroneous impression that little to no bias amplification has occurred as they aggregate positive and negative bias scores. Further, these metrics lack an ideal value, making them difficult to interpret. To address these shortcomings, we propose a new metric: \emph{Multi-Attribute Bias Amplification}. We validate our metric's utility through a bias amplification analysis on the COCO, imSitu, and CelebA datasets. Finally, we benchmark bias mitigation methods using our proposed metric, suggesting possible avenues for future bias mitigation efforts.
\end{abstract}

\section{Introduction}
Despite their intent to faithfully depict the world, visual datasets are undeniably subject to historical and representational biases~\cite{suresh2021framework,jo2020lessons}. Left unchecked, dataset biases are invariably learned by models, especially when they are sources of efficient features for supervised learning on a given dataset~\cite{gebru2018datasheets}. For example, an image captioning model can learn to generate gendered captions by exploiting contextual cues without ever ``looking'' at the person in the image~\cite{hendricks2018women}. Reliance on spurious correlations is undesirable since these learned associations do not always hold~\cite{sagawa2019distributionally,geirhos2020shortcut}. More significantly, these associations risk not only perpetuating harmful social biases but also \emph{amplifying} them~\cite{zhao2017mals}.

\begin{figure}[t]
\vskip 0.2in
    \centering
    \includegraphics[width=0.99\columnwidth]{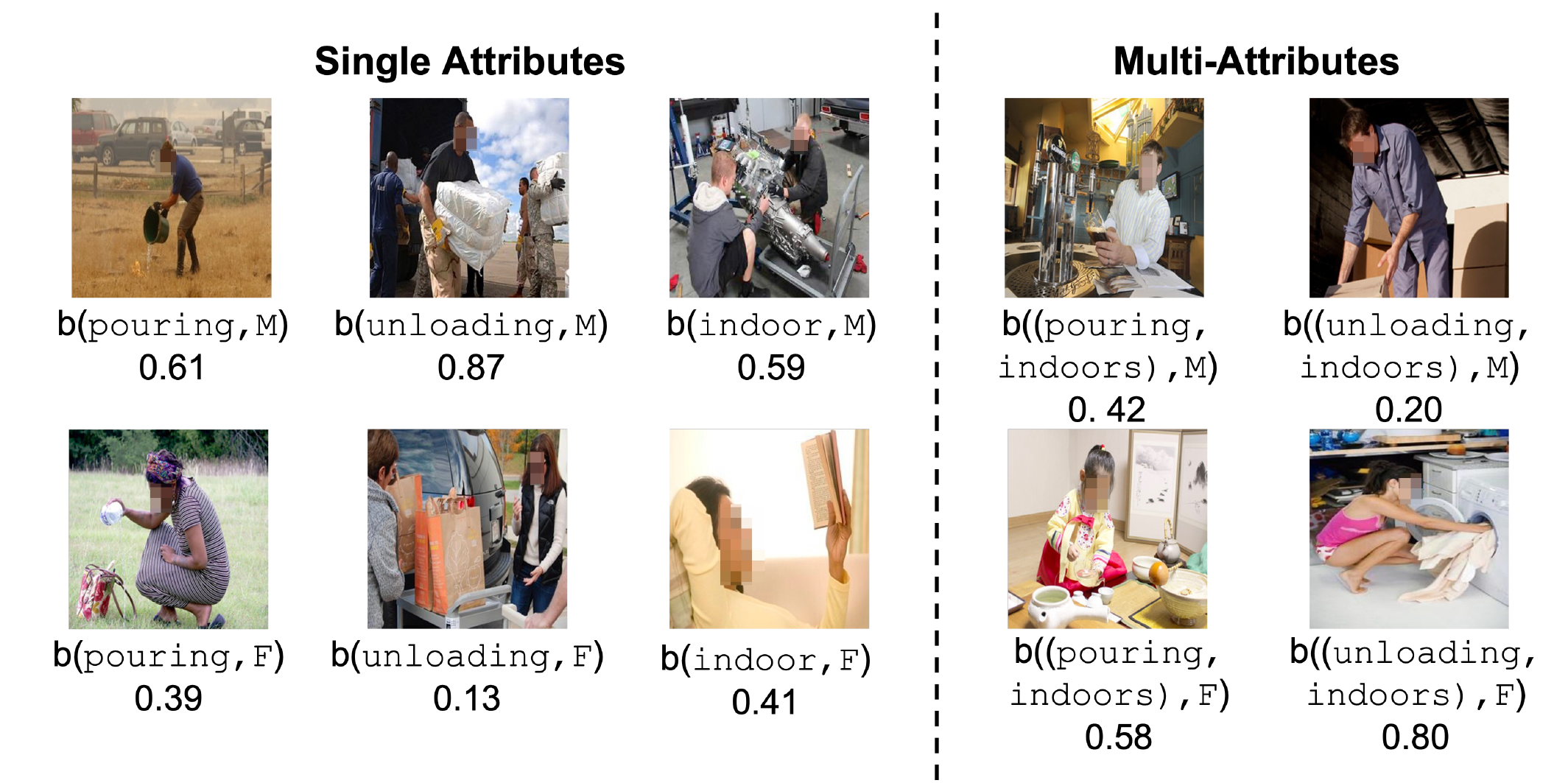}
    \alt{Two-column figure where the left-side shows the gender ratios for the single attributes that are skewed male and the right shows gender ratios for multiple attributes that are skewed female.}
    \caption{Bias scores (i.e., gender ratios) of the verbs \texttt{pouring} and \texttt{unloading} as well as location \texttt{indoors} in imSitu. While imSitu is skewed male (\texttt{M}) for the single attributes, the multi-attributes (e.g., \{\texttt{pouring}, \texttt{indoors}\}) are skewed female (\texttt{F}). Face pixelization is employed for privacy purposes.}
    \label{fig:fig1}
\vskip -0.2in
\end{figure}

The phenomenon of \emph{bias amplification} refers to when a model compounds the inherent biases of its training set at test time~\cite{zhao2017mals}. Bias amplification has been studied across many tasks~\cite{zhao2017mals,ramaswamy2020debiasing,wang2020fair,choi2020fair,jia2020mitigating,leino2018featurewise,wang2021biasamp,hirota2022quantifying,wang2019balanced,renduchintala2021gender}. Following \citet{zhao2017mals}, we focus on multi-label classification. While there are metrics~\cite{zhao2017mals,wang2021biasamp,wang2019balanced} that measure bias amplification in multi-label classification, they only consider the amplification between a single annotated attribute (e.g., \texttt{computer}) and a group (e.g., \texttt{female}). However, existing large-scale visual datasets often have multiple attributes per image (e.g., \{\texttt{computer}, \texttt{keyboard}\}). For example, $78.8\%$ of images in the Common Objects with Context (COCO)~\cite{lin2014microsoft} dataset contain more than one annotated attribute.  

More importantly, considering multiple attributes can reveal additional nuances not present when considering only single attributes. In imSitu~\cite{yatskar2016situation}, individually the verb \texttt{unloading} and location \texttt{indoors} are skewed \texttt{male} (Fig.~\ref{fig:fig1}). However, when considering \{\texttt{unloading, indoors}\} in conjunction, imSitu is skewed \texttt{female}. Significantly, men tend to be pictured unloading \emph{packages} outdoors whereas women are pictured unloading \emph{laundry} or \emph{dishes} indoors. Even when men are pictured indoors, they are unloading \emph{boxes} or \emph{equipment} as opposed to \emph{laundry} or \emph{dishes}. Models can similarly leverage correlations between a group and either single or multiple attributes simultaneously.

\smallsec{Multi-Attribute Bias Amplification Metric}
We propose \emph{multi-attribute bias amplification}, extending two previously proposed metrics \cite{zhao2017mals,wang2021biasamp}.\footnote{Our code is available at \url{https://github.com/SonyResearch/multi_bias_amp}.} Our new metric evaluates bias amplification arising from single and multiple attributes. We are the first to study multi-attribute bias amplification, highlighting that models exploit correlations between multiple attributes and group labels. We also address the issue that aggregated bias amplification metrics include summing positive and negative values. These values can cancel each other out, ostensibly presenting a smaller amount of amplification than what exists. Finally, as opposed to prior metrics which lack a clear ideal value, our metric is more interpretable.

\smallsec{Empirical Bias Amplification Analysis}
Using our proposed metric, we compare the performance of multi-label classifiers trained on COCO~\cite{lin2014microsoft}, imSitu, and CelebA~\cite{liu2015faceattributes}, standard benchmarks for bias amplification metrics~\cite{zhao2017mals,wang2019balanced,wang2021biasamp,hirota2022quantifying,ramaswamy2020debiasing}. As case studies, we consider bias amplification w.r.t. gender expression (COCO and imSitu), as well as hair color (CelebA) which represents the first bias amplification benchmark analysis for non-binary groups. We empirically demonstrate that on average bias amplification arising from single attributes is smaller than from multi-attributes. Thus, if we were only to consider individual attributes, not only would we obscure the nuance of understanding that multi-attributes provide but also potentially understate bias amplification.

\smallsec{Benchmarking Bias Mitigation Methods}
Finally, we benchmark different bias mitigation methods~\cite{zhao2017mals,wang2020fair,agarwal2020towards} on our metric and existing bias amplification metrics. \new{While prior works have demonstrated that models will learn to exploit different spurious correlations if one is mitigated~\cite{li2022whac, li2022discover}} we are the first to demonstrate that mitigation methods for single attribute bias can actually increase multi-attribute bias amplification. This further emphasizes the importance of our new metric as the magnitude of bias amplification is likely being underreported using single attribute metrics.

\section{Related Work}
\smallsec{Dataset Bias} 
Dataset bias is a well-studied problem in computer vision~\cite{torralba2011unbiased,devries2019everyone,buolamwini2018gender,zhu2014capturing,prabhu2020large}. Datasets are particularly predisposed to biases reflecting social inequities and disparities between true distributions and their digitized representations~\cite{suresh2021framework,jo2020lessons}. For example, datasets have been found to be demographically imbalanced~\cite{buolamwini2018gender,yang2020fair,zhao2021captionbias,dulhanty2019auditing,paullada2021data}, with a particular lack of representation of females and individuals with darker skin tones. Further, there are visual differences in the way individuals from different groups are represented or interact with objects in the images~\cite{wang2020revise,zhao2017mals,zhao2021captionbias}. In our work, we focus on how dataset biases are compounded by trained models.

\smallsec{Measuring Fairness} 
\new{A perfectly accurate model is unable to satisfy certain fairness metrics e.g., statistical parity~\cite{dwork2012fairness} as fairness constraints are not met in the ground-truth data~\cite{wang2021biasamp}.} In contrast, bias amplification metrics do not make recommendations as to the ideal underlying data distribution. By taking into account the correlations in the training set, bias amplification metrics instead capture additional biases introduced by a model. Bias amplification metrics can thus distinguish between cases when learned correlations are being over- or under-predicted, unlike fairness metrics such as TPR difference and FNR difference~\cite{hardt2016equality}.

\smallsec{Measuring Bias Amplification}
Beyond reproducing biases, machine learning models have also been found to amplify them~\cite{zhao2017mals}. \citet{hall2022systematic} empirically show that bias amplification is influenced by dataset bias, model capacity, and training schema. To quantify bias amplification, \citet{zhao2017mals} measure the difference in ground-truth object-group co-occurrences in the training set and  test set co-occurrences predicted by a model. Building on this, \citet{wang2021biasamp} propose a \emph{directional} bias amplifcation metric to disentangle bias arising from attribute prediction versus group prediction. Others focus on \emph{leakage}~\cite{wang2019balanced, hirota2022quantifying}, i.e., the change in a classifier's ability to predict group membership from ground-truth labels versus a model's predictions. Our work extends co-occurrence-based metrics~\cite{zhao2017mals,wang2021biasamp}. We are the first to consider how multi-attributes impact bias amplification and propose a novel metric to quantify this phenomenon.

\smallsec{Mitigating Bias Amplification}
Bias amplification mitigation methods focus on either the dataset or model. Dataset-level mitigation strategies tend to center on the use of generative adversarial networks (GANs)~\cite{ramaswamy2020debiasing,sattigeri2019fairness,sharmanska2020contrastive} and counterfactuals~\cite{kaushik2019learning,wang2021robustness} to augment training sets. 

More recent work~\cite{agarwal2022does} instead propose resampling strategies to address spurious correlations. Model-level strategies include corpus-level constraints~\cite{zhao2017mals}, adversarial debiasing~\cite{wang2019balanced}, and domain independent training~\cite{wang2020fair}. \new{Similar to previous works~\cite{shrestha2022investigation,zhao2022scaling}, we benchmark different mitigation methods. We demonstrate that existing strategies do not mitigate bias amplification from multiple attributes.}

\section{Multi-Attribute Bias Amplification}
In this section, we introduce our multi-attribute bias amplification metric. Throughout this paper, we scale all metrics by a factor of 100.

\smallsec{Identifying bias} 
\label{sec:notation}
We denote by $\mathcal{G}=\{g_1, \dots, g_t\}$ and ${\mathcal{A}=\{a_1,\dots,a_n\}}$ a set of $t$ group membership labels and $n$ attribute labels, resp. 
Let ${\mathcal{M}=\{m_1,\dots,m_{\ell}\}}$ denote a set of $\ell$ sets, containing all possible combinations of attributes, where $m_i$ is a set of attributes and $ \lvert m_i \rvert \in \{1,\dots, n\}$.\footnote{\new{For example, if $\mathcal{A}=\{a_1,a_2,a_3\}$, then $\mathcal{M}=\{\{a_1\}, \{a_2\}, \{a_3\}, \{a_1, a_2\}, \{a_1, a_3\}, \{a_2, a_3\}, \{a_1, a_2, a_3\} \}.$}} 
\new{Note $m\in\mathcal{M} \iff \text{co-occur}(m, g) \geq 1$ in both the ground-truth training set and test set, where $\text{co-occur}(m,g)$ is the number of times $m$ and $g$ co-occur.} We extend \citet{zhao2017mals}'s single-attribute bias score to a multi-attribute setting such that the bias score of $m\in\mathcal{M}$ w.r.t. $g\in\mathcal{G}$ is defined as:
\new{\begin{align}
\text{bias}_{\text{train}}(m,g)&=\frac{\text{co-occur}(m,g)}{ \sum_{g^\prime\in\mathcal{G}} \text{co-occur}(m,g^\prime)},
\end{align}
where $\text{co-occur}(m,g)$ denotes the number of times the combination of attributes $m$ and group membership label $g$ co-occur in the training set.}

\smallsec{Evaluating undirected multi-attribute bias amplification} 
We define our undirected multi-attribute bias amplification metric as: 
\begin{align}
    \text{Multi}_\text{MALS} &= X, \text{Var}(\Delta_{gm}) \label{eq:undirmultiba}
\end{align}
where
\new{\begin{align*}
X &= \frac{1}{\lvert \mathcal{M} \rvert}\sum_{g \in \mathcal{G}} \sum_{m \in \mathcal{M}}  \left\lvert\Delta_{gm}\right\rvert 
\end{align*}}
and
\begin{align*}
\begin{split}
\Delta_{gm} &=  \new{\mathbbm{1}\left[ \text{bias}_{\text{train}}(m, g) > \lvert
\mathcal{G}\rvert^{-1} \right] \cdot} \\ 
&\new{\left( \text{bias}_{\text{test}}(m, g) - \text{bias}_{\text{train}}(m, g)\right).}
\end{split}
\end{align*}
\new{Here $\text{bias}_{\text{test}}(m, g)$ denotes the bias score using the test set predictions of $m$ and $g$.} $\text{Multi}_\text{MALS}$ measures both the mean and variance over the change in bias score, $\Delta$, from the training set ground truths to test set predictions. By definition, $\text{Multi}_\text{MALS}$ only captures group membership labels that are positively correlated with a set of attributes\new{, i.e., due to the constraint that \new{$\text{bias}_{\text{train}}(m,g) > \lvert \mathcal{G}\rvert^{-1}$}}.

\smallsec{Evaluating directional multi-attribute bias amplification} Let $\hat{m}$ and $\hat{g}$ denote a model's prediction for attribute group, $m$, and group membership, $g$,
resp. \new{Further, without loss of generality, let $P_{\text{train}}(m=1)$ denote the probability of a training set sample containing the attribute set $m$, and $P_{\text{train}}(g=1)$ denote the probability of a training set sample containing the group membership label $g$.} We define our directional multi-attribute bias amplification metric as:

\begin{align}
    \text{Multi}_{\rightarrow} &= X, \text{Var}(\Delta_{gm}) \label{eq:dirmultiba}
\end{align}

where

\begin{align*}
    X &=\frac{1}{\lvert \mathcal{G} \rvert \lvert \mathcal{M} \rvert} \sum_{g\in \mathcal{G}}\sum_{m\in \mathcal{M}}  y_{gm}\left\lvert\Delta_{gm}\right\rvert+(1-y_{gm})\left\lvert-\Delta_{gm}\right\rvert,
\end{align*}
\begin{align*}
\begin{split}
    y_{gm}=\mathbbm{1} [ &\new{P_{\text{train}}(g=1, m=1)} \\ 
    &\new{> P_{\text{train}}(g=1) P_{\text{train}}(m=1) ]},
\end{split}
\end{align*}
and
\begin{align*}
    \Delta_{gm} &= \begin{cases}
    \new{P_{\text{test}}(\hat{m}=1 \vert g=1) - P_{\text{train}}(m=1 \vert g=1)} \\ \text{if measuring }G\rightarrow M\\
    \new{P_{\text{test}}(\hat{g}=1 \vert m=1) - P_{\text{train}}(g=1 \vert m=1)} \\
    \text{if measuring }M\rightarrow G
    \end{cases}
\end{align*}

Unlike $\text{Multi}_\text{MALS}$, $\text{Multi}_{\rightarrow}$ captures both positive and negative correlations, i.e., regardless of whether \new{$\text{bias}_{\text{train}}(m,g) > \lvert \mathcal{G}\rvert^{-1}$}. Moreover, \new{similar to the directional bias amplification metric proposed by \citet{wang2021biasamp}}, $\text{Multi}_{\rightarrow}$ takes into account group membership base rates and disentangles bias amplification arising from the group influencing the attribute(s) prediction (i.e., $\text{Multi}_{G \rightarrow M}$), versus amplification from the attribute(s) influencing the group prediction (i.e., $\text{Multi}_{M \rightarrow G}$). See Appendix~\ref{sec: app_relationship} for details on the relationship between the metrics.

\smallsec{Relation to existing single-attribute metrics} In Eqs.~(\ref{eq:undirmultiba}) and~(\ref{eq:dirmultiba}), if we only report $X$ with $\lvert \pm \Delta_{gm} \rvert = \sign(\pm \Delta_{gm})\lvert \pm \Delta_{gm} \rvert$ and $m_i \in \mathcal{M} \iff \lvert m_i \rvert =1$, then our metric $\text{Multi}_\text{MALS}$ reduces to~\citet{zhao2017mals}'s undirected bias amplification metric $\text{BiasAmp}_\text{MALS}$ and $\text{Multi}_{\rightarrow}$ reduces to~\citet{wang2021biasamp}'s directional bias amplification metric $\text{BiasAmp}_{\rightarrow}$. Refer to Appendix~\ref{sec:app_existing}  for formal definitions of $\text{BiasAmp}_\text{MALS}$ and $\text{BiasAmp}_{\rightarrow}$.

\section{Metric Advantages}
We compare our multi-attribute bias amplification metric with single-attribute metrics~\cite{zhao2017mals,wang2021biasamp}, using the MNIST dataset~\cite{deng2012mnist} to underscore the three main advantages of our metric: \new{(1) our metric accounts for co-occurrences with multiple attributes; (2) negative and positive values do not cancel each other out; and (3) our metric is more interpretable.}

\smallsec{Setup} We perform multi-label classification on a synthetically manipulated MNIST. For simplicity, we convert the task to binary classification~\cite{hall2022systematic} such that half of the classes are arbitrarily assigned to group ${g=0}$ or ${g=1}$. For the attributes, per image, we set a combination of three corner pixels ($a_1$: top left, $a_2$: bottom left, $a_3$: top right) to white. Thus, each image has a corresponding label $y = [g, a_1, a_2, a_3]$, where $a_i\in\{0,1\}$ corresponds to a pixel being colored black ($0$) or white ($1$). We train a LeNet-5~\cite{lecun1989handwritten} for $50$ epochs using SGD with batch size $32$, momentum $0.9$, and learning rate $10^{-3}$. We average over five random group assignments and report the 95\% confidence interval. 

\smallsec{Advantage 1: Our metric accounts for co-occurrences with multiple attributes} 
\label{sec:limit_multi}
If a model, for example, learns the combination of $a_1$ and $a_2$, denoted $\{a_1,a_2\}$, are correlated with $g$, it can exploit this correlation, potentially leading to bias amplification. By limiting the measurement to single attribute co-occurrences, $\text{BiasAmp}_\text{MALS}$ and $\text{BiasAmp}_{\rightarrow}$ do not account for amplification arising from co-occurrences with multiple attributes.

\begin{table}[t]
    \caption{Multi-attribute, $m$, bias scores w.r.t. group ${g = 1}$ using ground-truth, \new{$\text{bias}_{\text{train}}(m, g)$}, and predicted, \new{$\text{bias}_{\text{test}}(m, g)$}, labels. We report the $95\%$ confidence interval over five random assignments of $g$.}
    \label{tab:toy}
     \vskip 0.15in
    \scriptsize
     \centering
    \begin{tabular}{lrr}
        \toprule
         $m$&  \new{$\text{bias}_{\text{train}}(m, g)$} & \new{$\text{bias}_{\text{test}}(m, g)$} \\\midrule
         $\{a_1, a_2\}$ & $0.80$ & $0.92 \pm 0.1$\\
         $\{a_1, a_3\}$ & $0.49$ & $0.50 \pm 0.0$\\
         $\{a_2, a_3\}$ & $0.80$ & $0.99 \pm 0.0$\\
         $\{a_1, a_2, a_3\}$ & $0.94$ & $1.00 \pm 0.0$\\
         \bottomrule
    \end{tabular}
    \vskip -0.1in
\end{table}

To illustrate this, we manipulate MNIST so that the dataset is perfectly balanced w.r.t. to single attributes, i.e., \new{$\text{bias}_\text{train}(a_i, g) = 0.5$} ($\forall i\in[1,2,3]$), but skewed for multiple attributes. For example, although \new{$\text{bias}_\text{train}(a_1, g)=\text{bias}_\text{train}(a_2, g)=0.5$}, the bias score for the combination of $\{a_1, a_2\}$ and $g_1$ is \new{$\text{bias}_\text{train}(\{a_1,a_2\}, g_1)=0.8$}. Tbl.~\ref{tab:toy} shows the results using our trained models which achieve an mAP of $89.0 \pm 2.6$. Bias amplification is $0.0 \pm 0.0$ for all three single-attribute metrics. However, as shown in Tbl.~\ref{tab:toy}, the bias scores calculated w.r.t. multiple attributes has increased. Therefore, bias has been amplified but is not being captured by existing metrics. Significantly, by iterating over all $m\in\mathcal{M}$ our proposed metric accounts for amplification from both single attributes (i.e., $\lvert m\rvert =1$) and multiple attributes (i.e., $\lvert m\rvert >1$). Thus, we capture previously unidentified attributes exhibiting amplification. While existing metrics report amplification values close to 0, our multi-attribute metric returns $9.2 \pm 2.2$, $0.3 \pm 0.1$, $0.2 \pm 0.1$ for $\text{Multi}_\text{MALS}$, $\text{Multi}_{G \rightarrow M}$, and $\text{Multi}_{M \rightarrow G}$, resp.

\smallsec{Advantage 2: Negative and positive values do not cancel each other out}
Existing metrics aggregate over the difference in bias scores for each individual attribute. Suppose there is a dataset with two annotated attributes $a_1$ and $a_2$. It is possible that ${\Delta_{ga_1} \approx -\Delta_{ga_2}}$ for $\text{BiasAmp}_\text{MALS}$ or equivalently the difference in bias scores have opposite signs for $\text{BiasAmp}_\rightarrow$. In such cases, bias amplification would be approximately 0, which gives the impression little to no bias amplification has occurred. 

To give a concrete example, we arbitrarily set $a_1$, $a_2$, and $a_3$ in MNIST to $0$ with a probability of $0.7$, $0.2$, and $0.4$, resp. The model achieves an mAP of $85.2 \pm 9.9$. One of the models results in ${\text{BiasAmp}_\text{MALS}} \approx 0.0$, suggesting no bias amplification has occurred. However, upon closer inspection, the bias scores for individual attributes are $\Delta_{ga_1} = 0.61$ and $\Delta_{ga_2} = -0.60$. \citet{wang2021biasamp} recognize this limitation and suggest returning disaggregated results for each attribute-group pair. However, disaggregated values are difficult to interpret and make comparing models cumbersome if not infeasible. In contrast, our metric uses the absolute values of differences. Doing so ensures positive and negative bias amplifications over all attribute-group pairs do not cancel each other out. This allows us to report a single aggregated value, which is easier to interpret than disaggregated values per attribute-group pair.

\smallsec{Advantage 3: Our metric is more interpretable}
\label{sec:adv3}
There is a lack of intuition as to the \emph{ideal} bias amplification value. One interpretation is that smaller values are more desirable. This becomes less clear when values are negative, as occurs in several bias mitigation works~\cite{wang2020fair,ramaswamy2020debiasing}. Negative bias amplification indicates bias in the predictions is in the opposite direction than that in the training set. However, this is not always ideal. First, there often exists a trade-off between performance and smaller bias amplification values. Second, high magnitude negative bias amplification may lead to erasure of certain groups. For example, in imSitu, \new{$\text{bias}_{\text{train}}(\texttt{typing, F}) = 0.52$}. Negative bias amplification signifies that the model underpredicts (\texttt{typing, F}), which could reinforce negative gender stereotypes~\cite{zhao2017mals}.

Instead, we may want to minimize the distance between the bias amplification value and $0$. This interpretation offers the advantage that large negative values are also not desirable. However, a potential dilemma occurs when interpreting two values with the same magnitude but opposite signs, which is a value-laden decision and depends on the system's context. Additionally, under this alternative interpretation, Adv. 2 becomes more pressing as this suggests we are interpreting models as less biased than they are in practice.

Our proposed metric is easy to interpret. Since we use absolute differences, the ideal value is unambiguously 0. Further, reporting variance provides intuition as to whether amplification is uniform across all attribute-group pairs or if particular pairs are more amplified.

\section{Bias Amplification Analysis}
\label{sec:balance}
We now analyze the advantages of our proposed metric by evaluating bias amplification when group membership is balanced w.r.t. to single attributes.  

\subsection{Setup} In our experiments, we focus on COCO, imSitu, and CelebA, which contain multiple attributes per image and are frequently used in bias amplification analyses~\cite{zhao2017mals,wang2019balanced,ramaswamy2020debiasing}. First, for COCO, group membership is binary gender expression, i.e., $\{\texttt{female}\text{, }\texttt{male}\}$, and attributes correspond to objects. To obtain the group labels, we follow \citet{zhao2017mals} and use the provided captions. We only consider objects occurring ${>100}$ times with either group, leading to $52$ objects in total. Second, for imSitu, group membership is binary gender expression, and attributes correspond to verbs and location. We derive group labels from the gendered agent terms. We only consider the $361$ verbs that occur ${>5}$ with each group and have a binary location label (i.e., $\{\texttt{indoor}\text{, }\texttt{outdoor}\}$). Finally, for CelebA, group membership is non-binary hair color, i.e., $\{\texttt{black hair}\text{, }\texttt{ blond hair}\text{, }\texttt{ brown hair}\}$, and attributes correspond to other annotated physical characteristics in the dataset. We only consider the 23 physical characteristics that occur ${>500}$ times with each group.

To balance the datasets w.r.t. single attributes, $\forall (a,g) \in \mathcal{A} \times \mathcal{G}$ we greedily oversample, until the bias score \new{${\text{bias}_{\text{train}}(a, g) \in [|\mathcal{G}|^{-1} \pm \epsilon]}$}.\footnote{For COCO and imSitu, $\epsilon=0.025$. For CelebA, $\epsilon=0.07$, as we must balance across a larger number of groups.} We train ``balanced'' models on each of the three datasets using ResNet-50 architectures~\cite{he2016deep} initialized from weights learned on ImageNet~\cite{russakovsky2015imagenet}, where the classification layer has been replaced such that it jointly predicts group membership and attributes. Refer to Appendix~\ref{sec:app_preprocess} for more details.

\subsection{Experimental Analysis}
\label{sec:experimental_analysis1}
We analyze the different bias amplification metrics when using balanced models. First, we examine the effect of including multiple attributes. We then analyze the metrics when taking the absolute value versus the raw differences. Finally, we take a closer look into the variance of bias amplification scores per attribute.

\begin{table}[t!]
    
    \caption{We report multi-attribute bias amplification \new{with the variance in brackets} when varying $\lvert m_i \rvert$, the minimum number of attributes in a combination. $\lvert m_i \rvert \geq 1$ includes biases from single and multi-attributes. We report $95\%$ confidence interval over five models trained using random seeds for COCO (a), imSitu (b), and CelebA (c).}
    \label{tab:multi_compare}
      \vskip 0.15in
     \scriptsize
     \centering
    \begin{tabular}{lrr}
        \toprule
             (a) COCO &  $\lvert m_i \rvert \geq 2$ & $\lvert m_i \rvert  \geq 1$\\
        \midrule
            $\text{Multi}_{\text{MALS}}$ & $22.3 \pm 0.7$, [$4.6\pm 0.1$] & $21.9 \pm 0.2$, [$4.5 \pm 0.1$]\\
            $\text{Multi}_{M\rightarrow G}$ & $22.7 \pm 0.3$, [$12.9 \pm 0.2$] & $22.2 \pm 0.3$, [$13.0 \pm 0.0$]\\
            $\text{Multi}_{G\rightarrow M}$  & $0.3 \pm 0.0$, [$0.0 \pm 0.0$] & $0.3 \pm 0.0$, [$0.0 \pm 0.0$]\\
            \midrule
             (b) imSitu &  $\lvert m_i \rvert  \geq 2$ & $\lvert m_i \rvert  \geq 1$\\
            \midrule
            $\text{Multi}_{\text{MALS}}$ & $18.0 \pm 0.3$, [$3.0 \pm 0.1$] &  $9.4 \pm 0.2$, [$1.6 \pm 0.1$]\\
            $\text{Multi}_{M\rightarrow G}$ & $14.5 \pm 0.2$, [$4.1 \pm 0.2$] &  $13.0 \pm 0.1$, [$3.2 \pm 0.1$]\\
            $\text{Multi}_{G\rightarrow M}$  & $0.1 \pm 0.0$, [$0.0 \pm 0.0$] &  $0.1 \pm 0.0$, [$0.0 \pm 0.0$]\\
            \midrule
             (c) CelebA &  $\lvert m_i \rvert  \geq 2$ & $\lvert m_i \rvert  \geq 1$\\
            \midrule
            $\text{Multi}_{\text{MALS}}$ & $23.2 \pm 0.4$, [$2.3 \pm 0.1$] &  $23.1 \pm 0.4$, [$2.3 \pm 0.1$]\\
            $\text{Multi}_{M\rightarrow G}$ & $5.5 \pm 0.0$, [$0.0 \pm 0.0$] &  $5.5 \pm 0.0$, [$0.0 \pm. 0.0$]\\
            $\text{Multi}_{G\rightarrow M}$  & $0.6 \pm 0.0$, [$0.1 \pm 0.0$] &  $0.6 \pm 0.0$, [$0.1 \pm 0.0$]\\
        \bottomrule
        \end{tabular}
    \vskip -0.1in
\end{table}

\begin{table*}[t!]
    \scriptsize
    \caption{Performance, existing metrics, and our proposed multi-attribute metrics on balanced versions of COCO, imSitu and CelebA. There are three versions of the metrics: calculated using raw differences from training to test set (a), absolute value of differences (b), and the variances of these differences (c). We report the $95\%$ confidence interval over five models trained using random seeds.}
    \label{tab:breakdown}
    \vskip 0.15in
    \centering
        \begin{tabular}{p{0.12\textwidth}rrrrrrr}
        \toprule
             (a) Raw 
             &  mAP 
             & $\text{BiasAmp}_{\text{MALS}}$ 
             & $\text{Multi}_{\text{MALS}}$ 
             & $\text{BiasAmp}_{M\rightarrow G}$ 
             & $\text{Multi}_{M\rightarrow G}$ 
             & $\text{BiasAmp}_{G \rightarrow M}$ 
             & $\text{Multi}_{G\rightarrow M}$\\
        \midrule
             COCO & $53.8 \pm 0.1$ & $-1.9 \pm 0.2$ & $-16.3 \pm 0.2$ & $-1.5 \pm 0.2$ & $-1.4 \pm 0.4$ & $-0.0 \pm 0.0$ & $-0.1 \pm 0.0$\\
             imSitu & $67.0 \pm 0.1$ & $0.3 \pm 0.1$ & $-2.4 \pm 0.3$ & $0.7 \pm 0.1$ & $-0.5 \pm 0.1$ & $0.0 \pm 0.0$ & $0.0 \pm 0.0$\\
             CelebA & $78.2 \pm 0.1$ & $0.6 \pm 0.1$ & $-8.4 \pm 0.6$ & $-0.3\pm 0.0$ & $0.5 \pm 0.1$ & $0.6 \pm 0.0$ & $0.0 \pm 0.0$ \\
        \midrule
             (b) Absolute &  & $\text{BiasAmp}_{\text{MALS}}$ & $\text{Multi}_{\text{MALS}}$ & $\text{BiasAmp}_{M\rightarrow G}$ & $\text{Multi}_{M\rightarrow G}$ & $\text{BiasAmp}_{G \rightarrow M}$ & $\text{Multi}_{G\rightarrow M}$\\
        \midrule
             COCO && $5.5 \pm 0.2$ & $21.9 \pm 0.2$ & $5.0 \pm 0.2$ & $22.2 \pm 0.3$ & $0.3 \pm 0.0$  & $0.3 \pm 0.0$\\
             imSitu && $1.3 \pm 0.0$ & $9.4 \pm 0.2$ & $11.9 \pm 0.1$ & $13.0 \pm 0.1$ & $0.1 \pm 0.0$ & $0.1 \pm 0.0$\\
             CelebA && $27.4 \pm 1.1$ & $23.1 \pm 0.4$ & $0.8 \pm 0.0$ & $5.5 \pm 0.0$ & $1.1 \pm 0.0$  & $0.6 \pm 0.0$\\
        \midrule
             (c) Variance &              
             & $\text{BiasAmp}_{\text{MALS}}$ 
             & $\text{Multi}_{\text{MALS}}$ 
             & $\text{BiasAmp}_{M\rightarrow G}$ 
             & $\text{Multi}_{M\rightarrow G}$ 
             & $\text{BiasAmp}_{G \rightarrow M}$ 
             & $\text{Multi}_{G\rightarrow M}$\\
        \midrule
            COCO && $0.2 \pm 0.0$ & $4.5 \pm 0.1$ & $0.4 \pm 0.0$ & $13.0 \pm 0.2$ & $0.0 \pm 0.0$ & $0.0 \pm 0.0$\\
            imSitu && $0.1 \pm 0.0$ & $1.6 \pm 0.1$ & $2.5 \pm 0.0$ & $3.2 \pm 0.1$ & $0.0 \pm 0.0$ & $0.0 \pm 0.0$\\
            CelebA && $0.0 \pm 0.0$ & $2.7 \pm 0.1$ & $0.0 \pm 0.0$ & $2.2 \pm 0.0$ & $0.0 \pm 0.0$  & $0.0 \pm 0.0$\\
        \bottomrule
        \end{tabular}
        \label{tab:var}
    \vskip -0.1in
\end{table*}

\smallsec{More bias arises from multiple attributes} 
\label{sec:k}
To examine the effect of including multi-attributes, we vary the minimum number of attributes in a combination (i.e., $\lvert m \rvert$). In Tbl.~\ref{tab:multi_compare}, we consider all possible combinations of attributes for $\lvert m \rvert\geq 1$ (i.e., all attribute sets with at least a single attribute) or $\lvert m \rvert\geq 2$ (i.e., only attribute sets with at least two attributes). 

Across all datasets and metrics, the bias amplification when $\lvert m \rvert\geq 2$ is greater than or equal to that when $\lvert m \rvert\geq 1$, underscoring the importance of considering multiple attribute groups as detailed in Adv. 1. For COCO and imSitu, $\text{Multi}_{\text{MALS}}$ is greater when $\lvert m \rvert\geq 2$ than when $\lvert m \rvert\geq 1$, which implies that the mean bias amplification arising from single attributes is lower than that from multiple attributes. We note for CelebA, bias amplification is approximately equal when $\lvert m \rvert\geq 1$ and $\lvert m \rvert\geq 2$, indicating the amount of bias being amplified due to single attributes is negligible compared to the amount of bias being amplified due to multiple attributes (i.e., $\lvert m \rvert\geq 2$). 

 
When considering disaggregated values, there is considerably higher bias arising from certain singular verbs, such as \texttt{constructing}, \texttt{reading}, and \texttt{vacuuming}, for $M\rightarrow G$ prediction. Finally, we note for single attributes, larger bias amplification occurs with attributes that co-occur with males, but more for females when considering multiple attributes. See Appendix~\ref{sec:app_qual} for more qualitative results.

\smallsec{Raw differences collapse individual attributes}
Tbl.~\ref{tab:breakdown} shows how each metric differs when using raw or absolute differences. We find that the magnitude of mean bias amplification is significantly higher when using absolute differences versus raw differences. For example, on imSitu, $\text{BiasAmp}_{M\rightarrow G}$ is $0.7 \pm 0.1$ for raw differences, but when using absolute differences $\text{BiasAmp}_{M\rightarrow G}$ is $11.9 \pm 0.1$. Further, this effect is particularly evident for $G \rightarrow M$ where amplification appears to be approximately $0.0$ when in fact amplification does occur as evidenced by the absolute scores. This emphasizes the fact that negative and positive values for individual attributes are collapsed to zero when using raw differences, highlighting the advantage of summing over the absolute differences (see Adv. 2). Therefore, when using raw differences, we arrive at the following erroneous conclusion: models are minimally amplifying biases.

\smallsec{Bias amplification is not uniform across attribute groups} Considering the variance of the metrics, we observe the variance of multi-attribute $\text{Multi}_{\text{MALS}}$ is an order of magnitude higher than single-attribute $\text{BiasAmp}_{\text{MALS}}$. We find that this occurs as there is a small set of multi-attributes groups where more bias amplification is arising \new{(see Appendix~\ref{sec:app_bs} for bias score distribution plots and Appendix \ref{sec:app_qual} for qualitative analysis of top attributes that contribute to bias amplification).} In contrast, bias arising from single attributes is more uniform. As discussed in Adv. 3, having this insight can help guide potential interventions such that we can target the more problematic groups of attributes.

\subsection{Understanding Factors of Bias Amplification}
We examine three factors that can influence the magnitude of bias amplified, i.e., group salience, attribute group size, and model performance.

\smallsec{$G \rightarrow M$ bias amplification depends on person salience} Amplification arising from group to attribute prediction ($\text{BiasAmp}_{G \rightarrow M}$, $\text{Multi}_{G\rightarrow M}$) is consistently low for COCO and imSitu. Although spurious correlations with gender exist in many parts of the image~\cite{meister2022artifacts}, the person is the main source of gender cues. For datasets such as COCO and imSitu where the person may not be the image's focal point, it is likely group membership is less easily inferred and thus has a smaller impact on attribute prediction. To validate our intuition, we evaluate $\text{BiasAmp}_{G \rightarrow M}$ and $\text{Multi}_{G\rightarrow M}$ with images from COCO containing person bounding boxes of varying sizes. There is a strong positive correlation between bias amplification and the size of the person bounding box---i.e., Pearson's~$r$ of $0.89$ and $0.94$ for $\text{BiasAmp}_{G \rightarrow M}$ and $\text{Multi}_{G\rightarrow M}$, resp. This is in line with \citet{hall2022systematic}'s findings, which suggest that harder to recognize groups can result in lower bias amplification. In contrast, for CelebA where the person is the focal point of the image, we find group to attribute prediction is larger ($1.1 \pm 0.0$ for $\text{BiasAmp}_{G \rightarrow M}$ and $0.6 \pm 0.0$ for $\text{Multi}_{G\rightarrow M}$).

\begin{figure*}[t!]
\vskip 0.2in
    \centering
    \includegraphics[width=\textwidth]{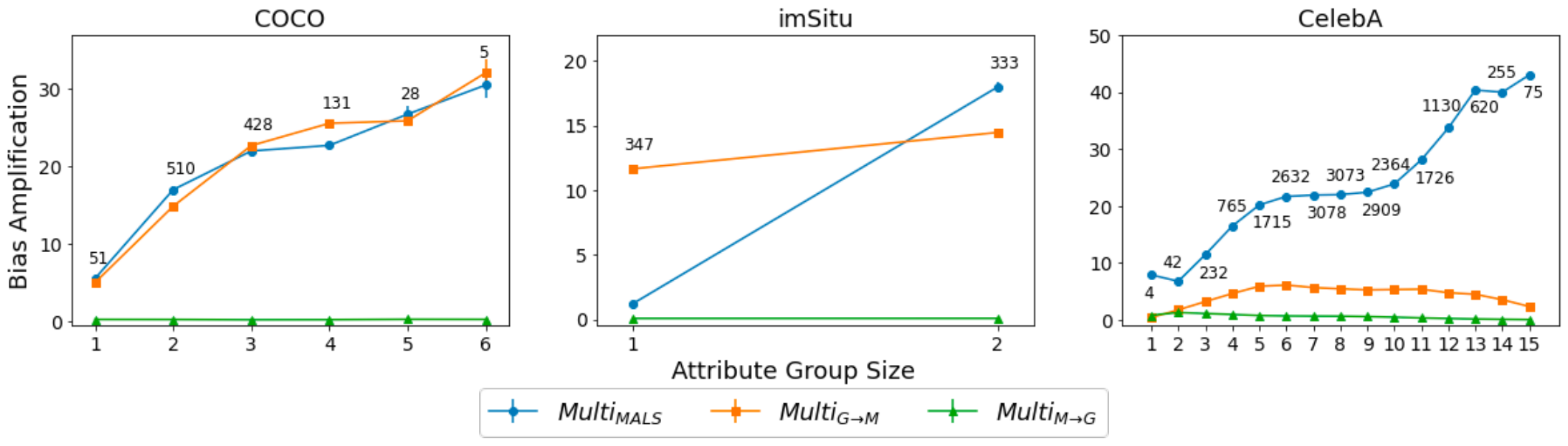}
    \alt{Three line charts showing bias amplification for COCO, imSitu, and CelebA for different attribute group sizes. Bias amplification increases as the attribute group size increases.}
    \caption{Visualization of bias amplification when varying the size of the attribute group, $\lvert m \rvert$. We plot the mean amplification score over five runs with random seeds of the model. Above each point, we include the number of attributes with size $\lvert m \rvert$. Error bars represent the standard deviation over runs.}
    \label{fig:vary_k}
\vskip -0.2in
\end{figure*}

\smallsec{Amplification occurs across all attribute group sizes}
While previous bias amplification metrics only consider amplification at $\lvert m_i \rvert = 1$, we provide results when varying attribute group sizes (see Fig.~\ref{fig:vary_k}). Across all attribute group sizes, we observe that amplification occurs. Further, for $\text{Multi}_\text{MALS}$ on all datasets and $\text{Multi}_{G\rightarrow M}$ on COCO and imSitu, bias amplification increases as we increase $\lvert m_i \rvert$. Although, $\text{Multi}_{M \rightarrow G}$ is smaller in magnitude, indicating there is not much bias arising from the attributes on group prediction, amplification exists (i.e., $>0$) and remains constant across the various attribute group sizes $\lvert m_i \rvert$.

\new{\smallsec{Performance and bias amplification are negatively correlated}
We calculate two performance metrics: mAP, the average precision over all individual attributes, and mAP*, a modified version of mAP that takes into account the different combinations of attributes in the ground-truth training set. As shown in Tbl.~\ref {tab:maps}, we find generally there is a negative correlation between bias amplification and performance metrics. For example, the Pearson's $r$ between mAP* and the $\text{Multi}_\text{MALS}$ are $-0.1 \pm 0.0$ and $-0.9 \pm 0.0$ for COCO for and CelebA respectively. This occurs since mAP* generally decreases as the attribute group size increases whereas amplification increases.
}

\begin{table}[t!]
    \centering
    \scriptsize
    \caption{\new{mAP*, a modified version of precision over all combinations of attributes, and the Pearson's $r$ between mAP* at various combination sizes with the respective multi-bias amplification metrics on balanced versions of COCO , imSitu, and CelebA. We do not report the correlations for imSitu as there are only two attribute group sizes.}}
    \vskip 0.15in
    \begin{tabular}{llll}
    \toprule
         & COCO & imSitu & CelebA \\
         \midrule 
         mAP* & $66.6 \pm 0.1$ & $31.4 \pm 0.0$ & $61.6 \pm 0.0$\\
         $\text{corr}(\text{mAP*}, \text{Multi}_\text{MALS})$ & $-0.1 \pm 0.0$ & - & $-0.9 \pm 0.0$\\
         $\text{corr}(\text{mAP*}, \text{Multi}_{M \rightarrow G})$ & $-0.5 \pm 0.1$ & - &  $1.0 \pm 0.0$\\
         $\text{corr}(\text{mAP*}, \text{Multi}_{G \rightarrow M})$ & $-0.2 \pm 0.0$ & - & $-0.4 \pm 0.0$\\
    \bottomrule
    \end{tabular}
    \label{tab:maps}
    \vskip -0.1in
\end{table}

\label{sec:experiments}
\section{Benchmarking Bias Mitigation Methods}
\label{sec:mitigate}

\begin{table*}[t!]
    \caption{Comparison of five mitigation methods---\textsc{Oversampling}, \textsc{RBA}~\cite{zhao2017mals}, \textsc{Adv}~\cite{wang2019balanced}, \textsc{DomInd}~\cite{wang2020fair}, and \textsc{Data Repair}~\cite{agarwal2022does}---against the baseline \textsc{Original}. We compare mAP, single attribute bias amplification, and multi-attribute bias amplification on COCO (a) and imSitu (b). We report the $95\%$ confidence interval over five models trained using random seeds. The bold values indicate the best performing method: distance to 0 for single-attribute and smallest value for multi-attribute metrics.}
    \label{tab:mitigate} 
    \vskip 0.15in
    \centering
    \scriptsize
    \centering
        \begin{tabular}{lrrrrrrr}
        \toprule
            (a) COCO &  mAP & $\text{BiasAmp}_{\text{MALS}}$ & $\text{Multi}_{\text{MALS}}$ & $\text{BiasAmp}_{M\rightarrow G}$ & $\text{Multi}_{M\rightarrow G}$ & $\text{BiasAmp}_{G\rightarrow M}$ & $\text{Multi}_{G\rightarrow M}$ \\\midrule
             \textsc{Original} & $53.4 \pm 0.2$ & $-0.6 \pm 0.3$ & $14.5\pm 0.6$ & $2.2 \pm 0.4$ & $12.5 \pm 0.2$ & $\mathbf{-0.0 \pm 0.0}$  & $0.4 \pm 0.0$\\
             \textsc{Oversampling} & $51.5 \pm 0.1$ & $1.1 \pm 0.1$ & $14.0 \pm 0.4$ & $-3.4 \pm 0.2$ & $12.5 \pm 0.3$ & $-0.2 \pm 0.0$ & $\mathbf{0.3 \pm 0.0}$\\
             \textsc{RBA} & $50.7 \pm 1.1$ & $3.8 \pm 1.7$ & $14.9 \pm 1.1$ & $-6.3 \pm 3.5$ & $17.3 \pm 2.2$ & $0.1 \pm 01$  & $0.4 \pm 0.0$\\
             \textsc{Adv} & $\mathbf{59.0 \pm 0.1}$ & $-0.7 \pm 0.9$ & $17.1 \pm 0.4$ & $7.0 \pm 0.6$ & $14.7 \pm 0.6$ & $0.1 \pm 0.0$  & $0.3 \pm 0.0$\\
             \textsc{DomInd} & $56.1 \pm 0.3$ & $0.4 \pm 0.6$ & $\mathbf{12.6 \pm 0.8}$ & $\mathbf{0.0 \pm 0.0}$ & $\mathbf{0.0 \pm 0.0}$ & $0.3 \pm 0.0$  & $0.3 \pm 0.0$\\
             \textsc{Data Repair} & $48.5 \pm 0.1$ & $\mathbf{0.3 \pm 0.1}$ & $17.2 \pm 0.3$ & $1.9 \pm 0.3$ & $11.7 \pm 0.2$ & $-0.0 \pm 0.0$  & $0.4 \pm 0.0$\\
    \midrule
            (b) imSitu &  mAP & $\text{BiasAmp}_{\text{MALS}}$ & $\text{Multi}_{\text{MALS}}$ & $\text{BiasAmp}_{M\rightarrow G}$ & $\text{Multi}_{M\rightarrow G}$ & $\text{BiasAmp}_{G\rightarrow M}$  & $\text{Multi}_{G\rightarrow M}$ \\\midrule
             \textsc{Original} & $67.1 \pm 0.1$ & $2.5 \pm 0.1$ & $37.5 \pm 0.1$ & $-0.3 \pm 0.1$ & $20.6 \pm 0.1$ & $0.0 \pm 0.0$  & $0.2 \pm 0.0$\\
             \textsc{Oversampling} & $66.3 \pm 0.1$ & $-4.5 \pm 0.2$ & $35.8 \pm 0.1$ & $-2.4 \pm 0.1$ & $20.1 \pm 0.1$ & $-0.0 \pm 0.0$ & $0.2 \pm 0.0$\\
             \textsc{RBA} & $54.7 \pm 0.5$ & $\mathbf{-1.4 \pm 0.3}$ & $35.4 \pm 0.3$ & $-6.2 \pm 0.3$ & $40.7 \pm 0.5$ & $-0.1 \pm 0.0$ & $0.3 \pm 0.0$\\
             \textsc{Adv} & $58.1 \pm 0.1$ & $4.1 \pm 0.3$ & $38.7 \pm 0.3$ & $0.6 \pm 0.4$ & $28.1 \pm 0.3$ & $-0.0 \pm 0.0$ & $0.2 \pm 0.0$\\
             \textsc{DomInd} & $\mathbf{69.6 \pm 0.1}$ & $10.2 \pm 0.9$ & $37.5 \pm 0.4$ & $\mathbf{0.0 \pm 0.0}$ & $\mathbf{0.0 \pm 0.0}$ & $0.1 \pm 0.0$ & $0.2 \pm 0.0$\\
             \textsc{Data Repair} & $62.3 \pm 0.1$ & $-1.8 \pm 0.1$ & $\mathbf{16.2 \pm 0.1}$ & $-0.1 \pm 0.1$ & $24.2 \pm 0.1$ & $\mathbf{-0.0 \pm 0.0}$ & $\mathbf{0.1 \pm 0.0}$\\
             \bottomrule
        \end{tabular}
        \label{tab:all_mitigate}
    \vskip -0.1in
\end{table*}

We now benchmark the performance of previously proposed bias mitigation methods~\cite{wang2019balanced, zhao2017mals, wang2020fair, agarwal2022does} using our multi-attribute bias amplification metrics.  

\subsection{Setup}
\smallsec{Datasets} 
As the mitigation methods we benchmark were originally proposed for gender bias mitigation, we focus on COCO and imSitu datasets. See Appendix~\ref{sec:app_mitigation} for results on CelebA. We obtain gender and attribute labels using the same process as described in Sec.~\ref{sec:balance}. The key difference here is that we do \emph{not} oversample to balance attribute-group co-occurrence pairs. As a result, for COCO, we have $18{,}177$, $4{,}545$, and $10{,}795$ images for the train, validation, and test splits, where $30.9\%$ of the instances are labeled female. For imSitu, we have $10{,}240$, $6{,}175$ and $24{,}698$ images for the train, validation, and test splits, where $40.7\%$ of the instances are labeled female.

\smallsec{Mitigation Methods}
Following \citet{wang2020fair}, we benchmark five mitigation methods. \new{As a simple baseline, we use oversampling, the method from Sec.~\ref{sec:balance}, which greedily samples to balance w.r.t. single attributes. We also select four popular strategies that have been evaluated for single-attribute bias amplification metrics: corpus constraints (\textsc{RBA})~\cite{zhao2017mals}, adversarial de-biasing (\textsc{Adv})~\cite{wang2019balanced}, data repair~\cite{agarwal2022does}\footnote{For \textsc{Data Repair}, the models are trained on a smaller number of instances since the method involves subsampling the dataset.}, and domain independent training (\textsc{DomInd})~\cite{wang2020fair}.\footnote{\citet{wang2020fair} propose four inference methods: conditioning on the known group, choosing the maximum over the groups, summing the probability of different groups, and summing the scores of the different groups. We report results on the first as it minimizes bias amplification in their work. See Appendix~\ref{sec:app_mitigation} for results on the other methods.}}

For each mitigation method, we set all parameters as proposed in respective papers. As a baseline, we train a ResNet-50 without any mitigation techniques using the same training protocol from Sec~\ref{sec:balance}. We refer to this model as \textsc{Original}.

\subsection{Experimental Analysis}
We now analyze the performance of different mitigation methods. Here we start by comparing their performance on single-attribute versus multi-attribute metrics. We conclude by discussing the trade-offs between different methods. All results are in Tbl.~\ref{tab:all_mitigate}.

\smallsec{Mitigating single-attribute bias amplification is not enough}
Even when mitigation methods decrease bias amplification for single attributes, this does not always translate to the multi-attribute case. For example, on COCO, while \textsc{Data Repair} outperforms all methods on $\text{BiasAmp}_{\text{MALS}}$, reducing bias amplification from $-0.6 \pm 0.3$ on \textsc{Original} to $0.3 \pm 0.1$, it increases $\text{Multi}_{\text{MALS}}$ from $14.5\pm 0.6$ to $17.2 \pm 0.3$. In fact, \textsc{DomInd} on COCO is the only method able to successfully mitigate amplification across all three multi-attribute metrics. Therefore, although current mitigation methods may ``work'' for single attributes, they increase multi-attribute bias amplification. Most significantly, this demonstrates that the overall amount of bias amplification may not actually be decreasing. Our finding underscores the need for mitigation methods that address bias amplification for both single and multi-attributes.

\smallsec{The best mitigation method is dataset dependent} Across the datasets, no mitigation method clearly outperforms another. For example, \textsc{DomInd} outperforms \textsc{Original} for COCO on all metrics except $\text{BiasAmp}_{G\rightarrow M}$ and $\text{Multi}_{G\rightarrow M}$; however, the method fares worse on imSitu. This is likely because \textsc{DomInd} attempts to distinguish between group membership within attributes (e.g., woman with computer versus man with computer). Since imSitu attributes (e.g., indoor or outdoor location) have more diverse appearances compared to objects in COCO, it may be more difficult to learn these boundaries. A potential avenue for inquiry is developing training methods that work well for more general attributes like those found in imSitu.

\smallsec{The best mitigation method is metric dependent} The relative ordering of the mitigation methods changes across metrics. For example, \textsc{DomInd} fares well for $M \rightarrow G$, consistently achieving bias amplification of $0.0 
\pm 0.0$. For \textsc{DomInd}, the prediction of attributes is conditioned on the group at test time, making it unlikely that the attributes would affect group prediction. However, for $G \rightarrow M$ amplification, \textsc{DomInd} either increases or does not significantly change the amount of bias amplification. In contrast, we have the case as with \textsc{Data Repair} on imSitu where $\text{Multi}_{G \rightarrow M}$ and $\text{Multi}_\text{MALS}$ decrease but $\text{Multi}_{M \rightarrow G}$ increases w.r.t. the baseline. \textsc{Data Repair} is ill-suited to scenarios where group amplification is of particular concern. Thus, only benchmarking on one metric can lead to the wrong conclusion that bias has been mitigated when in fact it has not changed or even increased. Overall, these results underscore the importance of considering which type of bias amplification is most relevant when deciding upon a mitigation strategy as the best-suited method may vary.

\section{Discussion}
\subsection{Implications for Mitigation Strategies}
\citet{wang2019balanced} illustrated that balancing datasets only w.r.t. group membership is insufficient. We implement more sophisticated balancing strategies: balancing each group with individual attributes and using \citet{agarwal2020towards}'s proposed data repair method. While balancing can reduce bias from single attributes, it does not work for multiple attributes. Multi-attribute bias amplification highlights the futility of balancing datasets. In particular, as balancing each group with all possible combinations of attributes is likely to be infeasible given the large number of attributes represented in datasets.

A potential avenue is to augment training datasets with synthetic images. While prior works~\cite{sharmanska2020contrastive,ramaswamy2020debiasing,sattigeri2019fairness} use GANs to mitigate bias amplification, these efforts have focused on face-centric datasets, such as CelebA and Diversity in Faces~\cite{merler2019diversity}, as opposed to more complex real-world image scenes. Using generative methods we can manipulate scenes to change the attributes pictured or alter a model's perception of group membership.\footnote{We do not advocate visual manipulations of group membership, given the ethical concerns of changing people's appearances, in particular for data collected without informed consent.} This would permit us to account for both single and multi-attributes when balancing datasets.

\subsection{Multi-Attribute Bias Amplification Limitations}

\smallsec{Reliance on annotations} 
As with existing co-occurrence metrics~\cite{zhao2017mals, wang2021biasamp}, our proposed metric only measures bias amplification w.r.t. annotated attributes. While we capture multi-attribute bias amplification, we cannot account for amplification that occurs due to unlabeled attributes. 
In addition, due to the lack of self-reported demographic annotations, we rely on third-party judgement of group membership. We acknowledge that relying on proxy judgements reifies the incorrect notion that, e.g., gender identity can be visually inferred and reducing gender to a binary is a harmful practice~\cite{hamidi2018gender,keyes2018agr}. Such reliance is a problem that many fairness researchers face~\cite{zhao2021captionbias,buolamwini2018gender,wilson} due to a lack of self-reported demographic information. 

\smallsec{Entanglement between metrics} Taking the absolute value of differences conflates two values of the same magnitude but with different signs. Both values are equal contributors to bias amplification. Some may argue that a bias amplification score of $-c$ is more desirable than $+c$ since ``bias'' has decreased. The preference between a positive or negative amplification is a value-laden decision that ultimately depends on the attribute and group. To illustrate, revisiting the example from Sec.~\ref{sec:adv3}, negative bias amplification for \{\texttt{typing, F}\} can contribute to erasure. Conversely, positive bias amplification for \{\texttt{baking, F}\} (which has a positive bias score of $0.6$ in imSitu's train set) reproduces and amplifies a harmful social stereotype. In both cases, regardless of the direction, bias amplification is undesirable.

Rather than making an overarching prescription, we propose using a multiplicity of fairness metrics. For example, reporting both raw and absolute differences permits a finer-grained analysis of where amplification is arising. 

\smallsec{Equal weighting of attribute groups}
Similar to single-attribute metrics, we weight all attribute combinations equally. However, depending on the context, one may want to place more weight on socially salient attribute combinations. For example, if a classifier trained on COCO is deployed in an athletics setting, amplification w.r.t. sports attributes such as \texttt{frisbee} or \texttt{tennis racket} may be more important. Alternatively, if the classifier is deployed in a classroom setting, these attributes may no longer be significant. We suggest metric users consider the deployment context of their system and adjust the weights accordingly.

\section{Conclusion}
Our proposed metric, Multi-Attribute Bias Amplification, illustrates the need to consider multiple attributes when measuring bias amplification. For perfectly ``balanced'' datasets, we find bias amplification occurs wrt multi-attributes, regardless of whether we use raw or absolute differences. Further, we are the first to show that methods which mitigate single attribute bias can inadvertently increase multi-attribute bias amplification. Overall, multi-attribute bias amplification provides a better characterization of the extent to which a model introduces bias from training to prediction.

\smallsec{Acknowledgements}
This work was funded by Sony Research. We thank William Thong and Julienne LaChance for their helpful comments and suggestions. 

\bibliography{ICML/icml_bib}
\bibliographystyle{icml2023}

\newpage
\appendix
\onecolumn
\section{Existing metrics definition} 
\label{sec:app_existing}
Using the notation from Sec.~\ref{sec:notation}, we provide formal definitions for $\text{BiasAmp}_{\text{MALS}}$ and $\text{BiasAmp}_\rightarrow$. We first provide the definition for undirected bias amplification~\cite{zhao2017mals}:
\begin{equation}
\new{\text{BiasAmp}_\text{MALS} = \frac{1}{\vert \mathcal{A} \vert}\displaystyle 
\sum_{g \in \mathcal{G}} \sum_{a \in \mathcal{A}} \mathbbm{1}\left[ \text{bias}_{\text{train}}(a, g) > \lvert
\mathcal{G}\rvert^{-1} \right] \cdot \left(
\text{bias}_{\text{test}}(a, g) - \text{bias}_{\text{train}}(a, g)\right).}
\end{equation} 
Note \new{$\text{bias}_{\text{test}}(a, g)$} is the bias score from the attribute and group label test set predictions, whereas \new{$\text{bias}_{\text{train}}(a, g)$} is the bias score from the attribute and group label training set ground truths.

\new{Let $\hat{a}$ and $\hat{g}$ denote a model's prediction for attribute, $a$, and group membership, $g$,
resp. Further, without loss of generality, let $P_{\text{train}}(a=1)$ denote the probability of a training set sample containing the attribute $a$, and $P_{\text{train}}(g=1)$ denote the probability of a training set sample containing the group membership label $g$. We now} provide the definition for directional bias amplification~\cite{wang2021biasamp}:

\begin{equation}
    \text{BiasAmp}_{\rightarrow} =\frac{1}{\lvert \mathcal{G} \rvert \lvert \mathcal{A} \rvert} \new{\sum_{g \in \mathcal{G}} \sum_{a \in \mathcal{A}}} y_{ga}\Delta_{ga}+(1-y_{ga})(-\Delta_{ga}),
    \end{equation}
where
\begin{equation*}
    \new{y_{ga}=\mathbbm{1}\left[ P_{\text{train}}(g=1, a=1) > P_{\text{train}}(a=1) P_{\text{train}}(g=1)\right]}
\end{equation*}
\begin{equation}
    \Delta_{ga} = \begin{cases}
    \new{P_{\text{test}}(\hat{a}=1 \vert g=1) - P_{\text{train}}(a=1 \vert g=1)}\\\text{if measuring }G\rightarrow A\\
    \new{P_{\text{test}}(\hat{g}=1 \vert a=1) - P_{\text{train}}(g=1 \vert a=1)}\\\text{if measuring }A\rightarrow G
    \end{cases}
\end{equation}

\section{Relationship between metrics} 
\label{sec: app_relationship}
We provide clarification on the relationship between the three multi-attribute bias amplification metrics. Importantly, we do not expect the metrics to be correlated. First, $\text{Multi}_\text{MALS}$ only captures positive correlations while directional metrics capture positive and negative. Second, the two directional metrics capture different phenomena: $M \rightarrow G$ measures the influence of attributes on group prediction and $G \rightarrow M$ measures the influence of group on attribute prediction.

\begin{figure}[b!]
\vskip 0.2in
    \centering
    \includegraphics[width=0.7\linewidth]{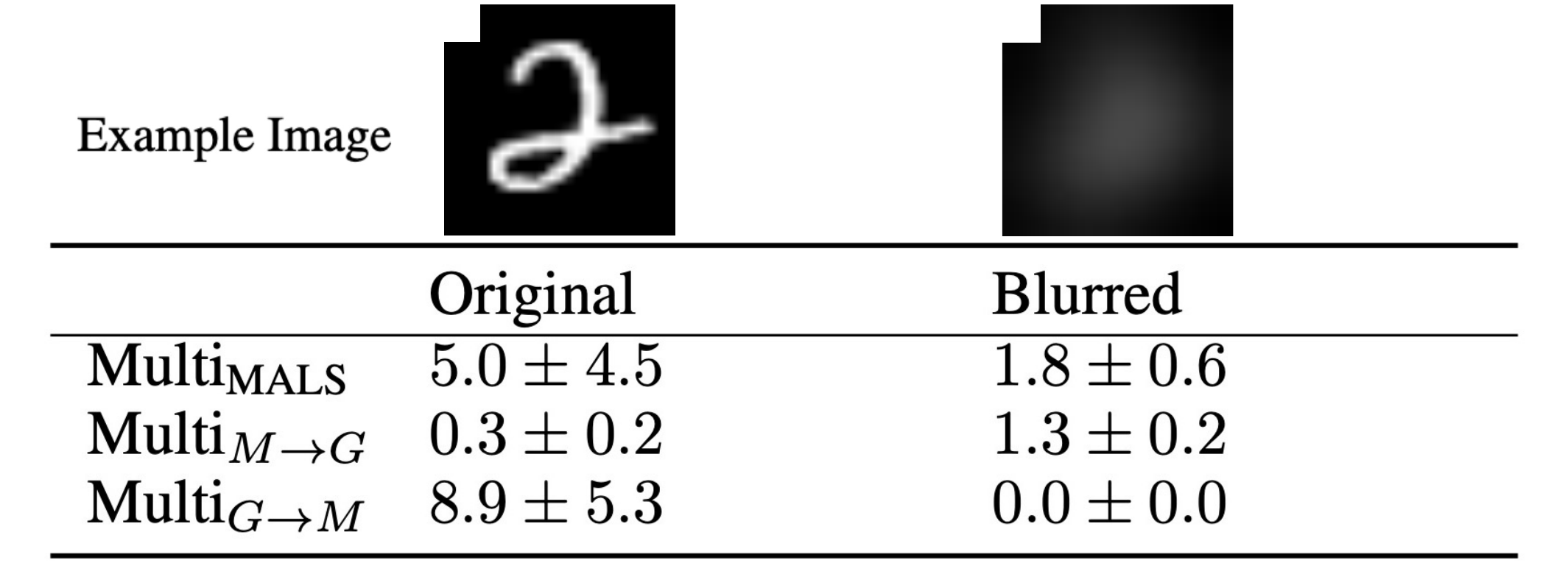}
    \alt{Table comparing the results of bias amplification on the original and blurred MNIST digit image. Bias amplification decreases for undirected bias amplification and group to attribute amplification but increases for attribute to group amplification.}
    \caption{Comparison of multi-attribute metrics for the original and blurred MNIST images. The example images have attributes $a_1 = 1$ and $a_2$, $a_3 = 0$.}
    \label{fig:reb}
    \vskip -0.2in
\end{figure}

To illustrate this, we use the MNIST setup from Sec.~\ref{sec:notation} but apply a Gaussian blur (radius 5), making group prediction hard. As shown in Fig.~\ref{fig:reb}, after blurring the digits, $\text{Multi}_\text{MALS}$and $\text{Multi}_{G \rightarrow M}$ decrease. $\text{Multi}_{M\rightarrow G}$ increases from 0.3 to 1.3. This is expected as the classifier relies on other attributes to predict the group. Conversely, since the group is difficult to recognize, the classifier is unlikely to rely on this feature when predicting attributes, causing $\text{Multi}_{G\rightarrow M}$ to decrease.

\section{Dataset preprocessing}
The greedy oversampling procedure results in $45{,}657$, $12{,}351$, and $27{,}499$ images respectively in the training, validation, and test sets for COCO. The splits for imSitu are $40{,}470$, $10{,}668$, and $17{,}036$ images. Finally, the splits for CelebA are $379{,}661$, $65{,}895$, and $52{,}397$ images.

The model is trained for 50 epochs using an Adam optimizer~\cite{kingma2015adam} with L2 weight decay of $10^{-6}$, batch size of $32$, and a learning rate of $10^{-5}$. Based on the known variance in fairness metrics~\cite{wang2021biasamp}, we train five models with random seeds and report the $95\%$ confidence interval.

\label{sec:app_preprocess}
We provide additional details on how the datasets are preprocessed.

\subsection{COCO}
\smallsec{Gender expression labels}
We follow \citet{zhao2021captionbias} and use an expanded word set to automatically derive perceived gender expression from the captions provided in the COCO~\cite{lin2014microsoft} dataset. When searching for gendered words, we first convert the captions to lowercase. We then use the following set of keywords to query the captions: 

\begin{itemize}
    \item Male: \{``male'', ``boy'', ``man'', ``gentleman'', ``boys'', ``men'', ``males'', ``gentlemen'', ``father'', ``boyfriend''\}
    \item Female: \{``female'', ``girl'', ``woman'', ``lady'', ``girls'', ``women'', ``females'', ``ladies'', ``mother'', ``girlfriend''\}
\end{itemize} 

One of the five COCO captions must contain a gendered term. Further, if both a male and female gendered term appear in the captions, the instance was discarded. This methodology matches that of prior works~\cite{zhao2017mals,wang2019balanced,agarwal2022does}

\smallsec{Attribute labels}
We only include attributes that have occurred more than 100 times with both male and female instances. This leaves us with the following 52 attributes: \{``person'', ``bicycle'', ``car'', ``motorcycle'', ``bus'', ``truck'', ``traffic light'', ``bench'', ``cat'', ``dog'', ``horse'', ``backpack'', ``umbrella'', ``handbag'', ``tie'', ``suitcase'', ``frisbee'', ``skis'', ``sports ball'', ``kite'', ``surfboard'', ``tennis racket'', ``bottle'', ``wine glass'', ``cup'', ``fork'', ``knife'', ``spoon'', ``bowl'', ``banana'', ``sandwich'', ``pizza'', ``donut'', ``cake'', ``chair'', ``couch'', ``potted plant'', ``bed'', ``dining table'', ``tv'', ``laptop'', ``remote'', ``cell phone'', ``microwave'', ``oven'', ``sink'', ``refrigerator'', ``book'', ``clock'', ``vase'', ``teddy bear'', ``toothbrush''\}. For attribute groups, we do not use ``person'' as this redundantly appears in all groups.

\subsection{imSitu}
\smallsec{Gender expression labels} 
To derive gender expression labels for imSitu, we use the agent annotations associated with each instance. We use the same set of keywords that we used for COCO to query male and female instances. 

\smallsec{Action and location labels} 
From the 504 annotated verbs in imSitu, we consider only those that co-occur with a gendered agent. Further, we only include actions that have occurred more than 5 times with both male and female instances. This leaves us with 361 verbs in total.

We turn location into a binary prediction between indoor and outdoor. To derive labels, we consider the top 50 place annotations in imSitu. We manually annotate each location as either ``indoor'' or ``outdoor,'' discarding locations that were ambiguous. This results in 25 location annotations that are divided into indoor and outdoor as follows:
\begin{itemize}
    \item Outdoors: \{``outdoors'', ``outside'', ``field'', ``street'', ``road'', ``sidewalk'', ``beach'', ``farm'', ``forest'', ``yard''\}
    \item Indoors: \{``room'', ``inside'', ``kitchen'', ``office'', ``gymnasium'', ``shop'', ``house'', ``hospital'', ``classroom'', ``workshop'', ``stage'', ``bed'', ``classroom'', ``living room'', ``bathroom''\}
\end{itemize}

\subsection{CelebA}
\smallsec{Hair color labels}
We use hair color as the group. To do so, we use three of the annotations for hair color included with CelebA: ``black hair,'' ``blond hair,'' ``brown hair.'' 

\smallsec{Physical attribute labels}
For the attributes in CelebA, we only include those that have occurred more than 500 times with each hair color respectively. This leaves us with the following 23 attributes: 
\{``Arched Eyebrows'',
``Bags Under Eyes'',
``Bangs'',
``Big Nose'',
``Eyeglasses'',
``Heavy Makeup'',
``High Cheekbones'',
``Male'',
``Mouth Slightly Open'',
``Narrow Eyes'',
``No Beard'',
``Oval Face'',
``Pale Skin'',
``Pointy Nose'',
``Receding Hairline'',
``Rosy Cheeks'',
``Smiling'',
``Straight Hair'',
``Wavy Hair'',
``Wearing Earrings'',
``Wearing Lipstick'',
``Wearing Necklace'',
``Young''\}.

\section{Training details} 
\label{sec:app_training}
All models in this work were developed using PyTorch. The models are trained and evaluated on 1 NVIDIA T4 Tensor Core GPU with 64 GB of GPU memory and 2.5 GHz Cascade Lake 24C processors. The operating system is Linux 64-bit Ubuntu 18.04. 

\section{Bias scores} 
\label{sec:app_bs}
\begin{figure}[t!]
\vskip 0.2in
    \centering
    \includegraphics[width=\textwidth]{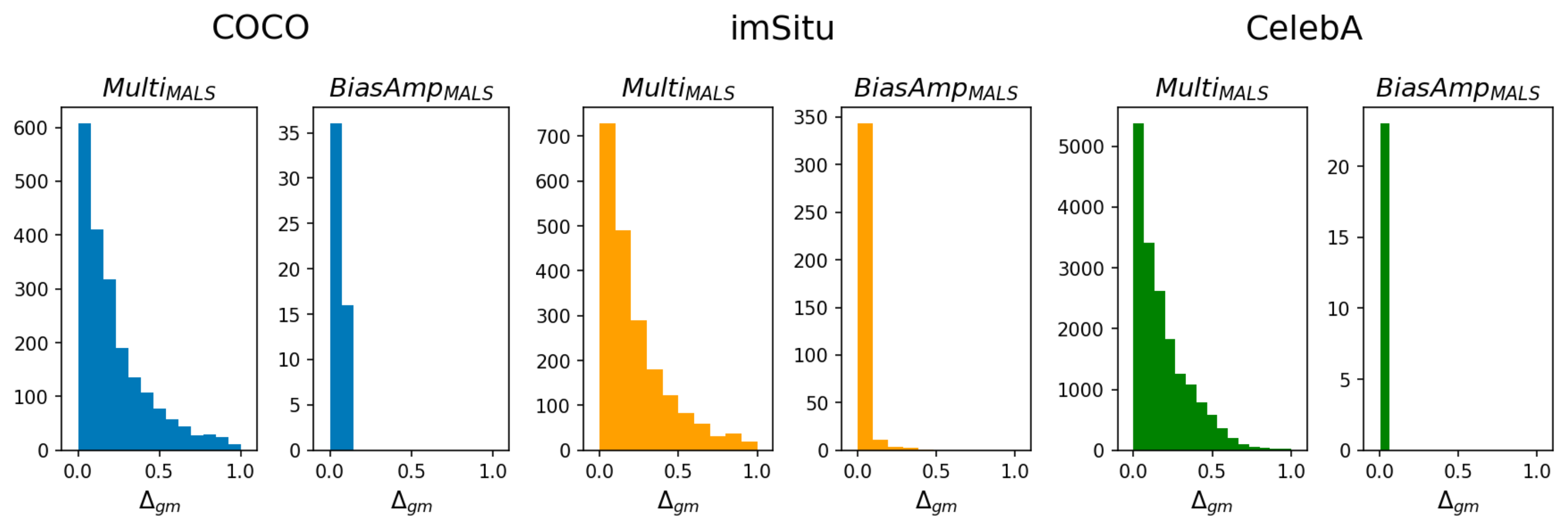}
    \alt{Histogram showing the change in bias scores for COCO, imSitu, and CelebA. The histograms have a right-tail with most instances occurring around 0.}
    \caption{Visualization of the absolute change in bias scores, $\lvert \Delta_{gm} \rvert$ from the ground-truth training set to the predicted labels distribution for undirected multiple-attribute and single-attribute bias amplification across COCO, imSitu, and CelebA.}
    \label{fig:bias_scores}
    \vskip -0.2in
\end{figure}

In Sec.~\ref{sec:experimental_analysis1}, we observe that the variance for undirected multi-attribute bias amplification is greater than that for undirected single attribute metrics. We visualize the disaggregated changes in bias score ($\Delta_{gm}$) for single and multi-attribute metrics across the three datasets (see Fig.~\ref{fig:bias_scores}). For all datasets, we see the distribution is right-skewed for $\text{Multi}_\text{MALS}$, meaning there is a small subset of multi-attribute groups where more amplification is occurring. 

\section{Bias mitigation methods}
\label{sec:app_mitigation}
\subsection{Method descriptions}
We provide a description of the five mitigation strategies used in Sec.~\ref{sec:mitigate}. We evaluate on the same test set consisting of $10{,}795$, $10{,}240$, and $11{,}351$ images for COCO, imSitu, and CelebA respectively.

\smallsec{Oversampling} We use a greedy oversampling process to balance group membership for each single attribute. \new{$\forall (a,g) \in \mathcal{A} \times \mathcal{G}$}, the sampling process terminates when the bias score \new{$\text{bias}_{\text{train}}(a, g) \in [|\mathcal{G}|^{-1} \pm \epsilon]$}, where $\epsilon=0.025$ for COCO, imSitu and $\epsilon=0.07$ for CelebA. This results in $45{,}657$, $40{,}470$, and $379{,}661$ training images for COCO and imSitu respectively. We train a ResNet-50 for 50 epochs using an Adam optimizer with batch size $32$, learning rate of $10^{-5}$, and weight decay of $10^{-6}$.

\smallsec{RBA} Reducing Bias Amplification~\cite{zhao2017mals} is employed after training. Here, the method uses corpus level constraints so that the predictions match a specified distribution. To ensure that the algorithm converges, the method uses Lagrangian relaxation. We use RBA to optimize the prediction distribution so it matches that of the training set.

\smallsec{Adversarial Debiasing} This method removes group information from the intermediate representation so that the model is not influenced by the group when making predictions. The goal during training is thus to improve the classifier's ability to predict attributes while making it difficult for the adversary to predict group membership. 

We remove group information from the final convolutional layer of a ResNet-50 (i.e., \textbf{adv @ conv5}. To train our adversarial debiasing method, we follow the procedure from~\citet{wang2019balanced}. We train with the adversarial loss using a learning rate of $5\times10^{-6}$ for $60$ epochs.

\smallsec{Domain Independent} As opposed to adversarial debiasing which promotes \textit{fairness through blindness}, \citet{wang2020fair} claim that domain independent training promotes \textit{fairness through awareness}. Concretely, domain independent training attempts to learn to differentiate between the same attribute for different groups. For example, in the case of gender bias amplification on COCO, the classifier will attempt to learn the boundary between (\texttt{computer, F}) and (\texttt{computer, M}). We train the ResNet-50 for 50 epochs using an Adam optimizer with batch size $32$, learning rate of $10^{-5}$, and weight decay of $10^{-6}$. 

\new{At inference time, there are $gn$ predictions (i.e., $n$ predictions for each group). To select which predictions we use, we take the predictions associated with the ground-truth group membership for the instance. We also provide results in Tbl.~\ref{tab:dom_ind_add} for the three other inference methods introduced by \citet{wang2020fair}.}

\begin{table*}[t!]
    \caption{Comparison of three inference methods for \textsc{DomInd}~\cite{wang2020fair}: (1) choosing the maximum over groups, (2) summing the probability of different groups, and (3) summing the sores of different groups. We compare mAP, single attribute bias, and multi-attribute bias amplification on COCO, imSitu, and CelebA. For notation, $y$, $g$, and $x$ refer to the attributes, group, and image respectively. $\lvert G \rvert$ indicates the number of groups.}
    \label{tab:dom_ind_add} 
    \vskip 0.15in
    \centering
    \scriptsize
    \centering
        \begin{tabular}{lrrrrrrr}
        \toprule
            COCO &  mAP & $\text{BiasAmp}_{\text{MALS}}$ & $\text{Multi}_{\text{MALS}}$ & $\text{BiasAmp}_{M\rightarrow G}$ & $\text{Multi}_{M\rightarrow G}$ & $\text{BiasAmp}_{G\rightarrow M}$ & $\text{Multi}_{G\rightarrow M}$ \\\midrule
             $\lvert G \rvert N \text{sigm, max } P(y\vert g, x)$ & $54.7$ & $-61.4$ & $59.7$ & $-3.1$ & $39.6$ & $-0.2$ & $0.3$ \\
             $\lvert G \rvert N \text{sigm } \sum_{g} P(y\vert g, x)$ & $54.7$ & $-8.4$ & $31.3$ & $-6.1$ & $31.2$ & $-0.1$ & $0.3$\\
             $\lvert G \rvert N \text{sigm } \sum_{g} s(y\vert g, x)$ & $54.8$ & $-61.4$ & $ 59.9$ & $-3.1$ & $39.6$ & $-0.1$ & $0.3$\\
        \midrule
            imSitu &  mAP & $\text{BiasAmp}_{\text{MALS}}$ & $\text{Multi}_{\text{MALS}}$ & $\text{BiasAmp}_{M\rightarrow G}$ & $\text{Multi}_{M\rightarrow G}$ & $\text{BiasAmp}_{G\rightarrow M}$ & $\text{Multi}_{G\rightarrow M}$ \\\midrule
            $\lvert G \rvert N \text{sigm, max } P(y\vert g, x)$ & $49.9$ & $-64.7$ & $59.7$ & $-3.1$ & $39.6$ & $-0.2$ & $0.3$\\
            $\lvert G \rvert N \text{sigm } \sum_{g} P(y\vert g, x)$ & $51.1$ & $-5.5$ & $36.3$ & $-5.0$ & $37.1$ & $-0.0$ & $0.3$\\
            $\lvert G \rvert N \text{sigm } \sum_{g} s(y\vert g, x)$ & $49.4$ & $-64.7$ & $32.8$ & $-13.0$ & $70.3$ & $-0.0$ & $0.5$\\
        \midrule
            CelebA &  mAP & $\text{BiasAmp}_{\text{MALS}}$ & $\text{Multi}_{\text{MALS}}$ & $\text{BiasAmp}_{M\rightarrow G}$ & $\text{Multi}_{M\rightarrow G}$ & $\text{BiasAmp}_{G\rightarrow M}$ & $\text{Multi}_{G\rightarrow M}$ \\\midrule
            $\lvert G \rvert N \text{sigm, max } P(y\vert g, x)$ & $37.4$ & $-99.0$ & $40.6$ & $-13.6$ & $34.2$ & $-3.9$ & $0.7$\\
            $\lvert G \rvert N \text{sigm } \sum_{g} P(y\vert g, x)$ & $ 38.0$ & $-3.6$ & $8.1$ & $-4.0$ & $27.5$ & $30.7$ & $0.8$\\
            $\lvert G \rvert N \text{sigm } \sum_{g} s(y\vert g, x)$ & $37.8$ & $-99.0$ & $-13.6$ & $-3.8$ & $43.2$ & $34.2$ & $0.7$\\
            \bottomrule
        \end{tabular}
    \vskip -0.1in
\end{table*}

\smallsec{Data Repair} \citet{agarwal2022does} attempt to repair the existing dataset using a fair selection process. Concretely, this method curates a dataset via subsampling until co-occurring attributes are well represented. In the supervised setting, the method uses a greedy approach that aims to minimize $c_v$, which is equal to standard deviation divided by the mean number of images per attribute. We use five random seeds, resulting in a mean coefficient variation of $1.78$, $5.83$, and $0.65$ for COCO, imSitu, and CelebA. The subsampled train and validation set size $3{,}000 \times g$ and $500 \times g$ where $g = 2$ for COCO, imSitu and $g = 3$ for CelebA.

\subsection{Additional mitigation results}
We report the result of applying the mitigation methods on the CelebA dataset in Tbl.~\ref{tab:app_mitigate}. It is important to note that the mitigation methods (e.g., \textsc{RBA}, \textsc{Adv}, \textsc{DomInd}, and \textsc{Data Repair}) were originally proposed for gender bias and only benchmarked on a binary group. Here, we extend the mitigation methods for the non-binary group of hair color. For single-attribute metrics, proposed methods can help mitigate bias amplification for specific metrics. For example, \textsc{Adv} reduces $\text{BiasAmp}_\text{MALS}$ from $2.0 \pm 0.3$ to $1.0 \pm 1.2$. However, overall, we see that mitigation methods are unable to reduce bias amplification, even increasing amplification in some instances, especially with respect to the multi-attribute metrics. 

\begin{table*}[ht!]
    \caption{Comparison of five mitigation methods---\textsc{Oversampling}, \textsc{RBA}~\cite{zhao2017mals}, \textsc{Adv}~\cite{wang2019balanced}, \textsc{DomInd}~\cite{wang2020fair}, and \textsc{Data Repair}~\cite{agarwal2022does}---against the baseline \textsc{Original}. We compare mAP, single attribute bias amplification, and multi-attribute bias amplification on CelebA. We report the $95\%$ confidence interval over five models trained using random seeds. The bold values indicate the best performing method: distance to 0 for single-attribute and smallest value for multi-attribute metrics.}
    \label{tab:app_mitigate} 
    \vskip 0.15in
    \centering
    \scriptsize
    \centering
        \begin{tabular}{lrrrrrrr}
        \toprule
            CelebA &  mAP & $\text{BiasAmp}_{\text{MALS}}$ & $\text{Multi}_{\text{MALS}}$ & $\text{BiasAmp}_{M\rightarrow G}$ & $\text{Multi}_{M\rightarrow G}$ & $\text{BiasAmp}_{G\rightarrow M}$ & $\text{Multi}_{G\rightarrow M}$ \\\midrule
             \textsc{Original} & $79.0 \pm 0.2$ & $2.0 \pm 0.3$ & $\mathbf{17.5 \pm 0.8}$ & $\mathbf{-0.1 \pm 0.1}$ & $2.3 \pm 0.2$ & $\mathbf{0.1 \pm 0.1}$  & $\mathbf{0.2 \pm 0.0}$\\
             \textsc{Oversampling} & $73.9 \pm 0.1$ & $-1.1 \pm 0.1$ & $19.2 \pm 0.5$ & $-0.9 \pm 0.0$ & $4.0 \pm 0.0$ & $-0.2 \pm 0.1$ & $0.5 \pm 0.0$\\
             \textsc{RBA} & $71.5 \pm 2.7$ & $25.8 \pm 12.9$ & $38.8 \pm 11.2$ & $-8.8 \pm 4.0$ & $21.0 \pm 10.5$ & $-0.8 \pm 0.3$  & $0.2 \pm 0.1$\\
             \textsc{Adv} & $73.9 \pm 0.1$ & $\mathbf{1.0 \pm 1.2}$ & $27.4 \pm 2.6$ & $4.9 \pm 0.8$ & $22.3 \pm 2.1$ & $-3.3 \pm 0.1$  & $0.4 \pm 0.0$\\
             \textsc{DomInd} & $\mathbf{82.7 \pm 0.1}$ & $3.6 \pm 0.3$ & $18.7 \pm 0.9$ & $-2.1 \pm 0.0$ & $\mathbf{0.0 \pm 0.0}$ & $0.3 \pm 0.1$  & $0.3 \pm 0.0$\\
             \textsc{Data Repair} & $82.2 \pm 0.2$ & $-10.6 \pm 0.3$ & $27.4 \pm 0.7$ & $-0.5 \pm 0.3$ & $3.3 \pm 0.2$ & $-0.5 \pm 0.1$  & $0.4 \pm 0.0$\\
             \bottomrule
        \end{tabular}
        \label{tab:imsitu_mitigate}
    \vskip -0.1in
\end{table*}

\begin{table}[t]
    \scriptsize
    \centering
    \caption{Top three groups of attributes with the largest contribution to bias amplification when training and evaluating on a perfectly ``balanced'' dataset and the difference in bias score for the respective attribute group. The ranking is calculated by averaging over the deltas with random seed assignments. We report the mean difference in bias score and the $95\%$ confidence interval. Bolded values indicate the bias is amplifying for male gender expression for COCO and imSitu. Bolded values indicate bias is amplifying for black hair color and italic values indicate bias is amplifying for blonde hair. }
        \vskip 0.15in
    \begin{tabular}{p{0.012\textwidth}p{0.2\textwidth}p{0.7\textwidth}}
    \toprule
         Rank & COCO Single & COCO Multi \\
         \midrule
          1 & \{\texttt{bus}\} & \textbf{\{\texttt{bus, banana}\}} \\
          & $14.6 \pm 0.8$ & $100.0 \pm 0.0$ \\
          2 & \{\texttt{dog}\} & \{\texttt{bottle,cup,cake,dining table,microwave,refrigerator}\} \\
          & $13.3 \pm 1.0$ & $100.0 \pm 0.0$\\
          3 & \textbf{\{\texttt{surfboard}\}} & \{\texttt{dog,pizza,couch}\} \\
          & $12.5 \pm 1.6$ & $100.0 \pm 0.0$\\ 
        \midrule 
        Rank & imSitu Single & imSitu Multi \\
        \midrule 
        1 &\textbf{\{\texttt{packaging}\}} &  \textbf{\{\texttt{indoor, crying}\}} \\
        & $48.3 \pm 0.0$ & $86.7 \pm 23.4$\\
        2 &\textbf{\{\texttt{giggling}\}} & \textbf{\{\texttt{indoors, drumming}\}} \\
        & $36.4\pm 3.7$ & $85.7 \pm 4.2$ \\
        3 &\{\texttt{confronting}\} & \textbf{\{\texttt{indoors, flinging}\}} \\
        & $30.2 \pm 2.0$ & $81.8 \pm 0.0$ \\
        \midrule 
        Rank & CelebA Single & CelebA Multi \\
        \midrule 
        1 &\{\texttt{receding hairline}\} &  \textit{\{\texttt{arched eyebrows, heavy makeup, no beard, pale skin, pointy nose, receding hairline, wearing earrings, wearing lipstick}\}} \\
        & $6.5 \pm 1.0$ & $100.0 \pm 0.0$\\
        2 & \textit{\{\texttt{rosy cheeks}\}} & \{\texttt{eyeglasses, high cheekbones, mouth slightly open, no beard, pointy nose, smiling, wavy hair, wearing earrings, wearing lipstick}\} \\
        & $3.9 \pm 0.1$ & $100.0 \pm 0.0$ \\
        3 &\textit{\{\texttt{narrow eyes}\}} & \textbf{\{\texttt{heavy makeup, high cheekbones, narrow eyes, no beard, receding hairline, smiling, straight hair, wearing lipstick, wearing necklace, young}\}} \\
        & $3.7 \pm 1.0$ & $100.0 \pm 0.0$ \\
    \bottomrule
    \end{tabular}
    \label{tab:balanced_qual}
        \vskip -0.1in
\end{table}

\begin{table*}[!htbp]
\scriptsize
    \centering
    \caption{Top three groups of attributes with the largest contribution to bias amplification for bias mitigation methods and the difference in bias score for the respective attribute group. Results are reported on COCO (top) and imSitu (bottom). The ranking is calculated by averaging over the deltas for five runs with random seed assignments. We report the mean difference in bias score and the $95\%$ confidence interval. Bolded values indicate the bias is amplifying for male gender expression.}
    \vskip 0.15in
    \bgroup
    \begin{tabular}{l p{0.13\textwidth}p{0.13\textwidth}p{0.15\textwidth}p{0.14\textwidth}p{0.14\textwidth}p{0.13\textwidth}}
        \toprule
        Rank & \textsc{Original} & \textsc{Oversample}  & \textsc{RBA} & \textsc{Adv} & \textsc{DomInd} & \textsc{Data Repair}\\ \midrule
        1 & \texttt{\{bench,dog,cup\}} & \texttt{\{bench,dog,cup\}} & \{\texttt{sport ball}, \texttt{potted plant\}} & \texttt{\textbf{\{bench,umbrella, potted plant\}}} & \texttt{\textbf{\{bench,umbrella, potted plant\}}} & \texttt{\textbf{\{sports ball,kite\}}}\\ 
        & $100.0 \pm 0.0$ & $100.0 \pm 0.0$ & $80.0 \pm 0.0$ & $72.8 \pm 3.9$ & $67.0 \pm 8.6$ & $100.0 \pm 0.0$\\
        2 & \texttt{\{sports ball,surfboard\}} & \texttt{\{sports ball,surfboard\}} & \textbf{\texttt{\{car,motorcycle, truck,handbag\}}} & \textbf{\texttt{\{car,motorcycle, truck,handbag\}}} & \texttt{\textbf{\{backpack,handbag, chair,cell phone\}}} & \texttt{\{car, surfboard\}}\\ 
        & $100.0 \pm 0.0$ & $100.0 \pm 0.0$ & $76.4 \pm 5.3$ & $67.1 \pm 5.3$ & $56.8 \pm 17.0$ & $83.3 \pm 0.0$\\
        3 & \texttt{\{sports ball,potted plant\}} & \textbf{\texttt{\{handbag,pizza, chair,dining table\}}} & \{\texttt{frisbee}, \texttt{clock}\} & \texttt{\{car,laptop\}} & \texttt{\textbf{\{car,motorcycle, truck,handbag\}}} & \texttt{\textbf{\{bus, cell phone\}}}\\ 
        & $80.0 \pm 0.0$ & $80.9 \pm 17.5$ & $76.4 \pm 5.3$ & $63.6 \pm 0.0$ & $56.6 \pm 9.8$ & $81.3 \pm 0.0$\\
        \midrule
        Rank & \textsc{Original} & \textsc{Oversample} & \textsc{RBA} & \textsc{Adv} & \textsc{DomInd} & \textsc{Data Repair}\\ \midrule 
        1 & \textbf{\{\texttt{indoors, crying}\}} & \textbf{\{\texttt{indoors, crying}\}} & \textbf{\{\texttt{indoors, crying}\}} & \textbf{\{\texttt{indoors, repairing}\}} & \textbf{\{\texttt{indoors, repairing}\}} & \{\texttt{indoors, slicing}\}\\
        & $100.0 \pm 0.0$ & $100.0 \pm 0.0$ & $100.0 \pm 0.0$ & $78.5 \pm 2.7$ & $64.0 \pm 28.0$ & $100.0 \pm 0.0$\\
        2 & \textbf{\texttt{\{indoor, repairing\}}} & \textbf{{\texttt{indoors, drumming}\}}} & \textbf{\texttt{\{indoor, repairing\}}} & \{\texttt{indoors, checking}\} & \textbf{\{\texttt{indoors, assembling}\}} & \textbf{\{\texttt{indoors, watering}\}}\\
        & $80.0 \pm 0.0$ & $100.0 \pm 0.0$ & $80.0 \pm 0.0$ & $64.1 \pm 8.3$ & $55.6 \pm 0.0$ & $100.0 \pm 0.0$\\
        3 & \textbf{{\texttt{indoors, drumming}\}}} & \textbf{\{\texttt{indoors, resting}\}}  & \textbf{\{\texttt{indoors, resting}\}} & \{\texttt{checking}\} & \textbf{\texttt{\{indoor, racing\}}} & \textbf{\{\texttt{indoors, pumping}\}}\\
        & $80.0 \pm 35.1$ & $67.1 \pm 4.7$ & $61.4 \pm 13.4$ & $61.7 \pm 7.2$ & $55.0 \pm 21.5$ & $100.0 \pm 0.0$\\
        \bottomrule
    \end{tabular}
    \egroup
    \label{tab:group_qual}
        \vskip -0.1in
\end{table*}

\begin{table*}[!htbp]
\scriptsize
    \centering
    \caption{Top three groups of attributes with the largest contribution to bias amplification for bias mitigation methods on CelebA and the difference in bias score for the respective attribute group. The ranking is calculated by averaging over the deltas for five runs with random seed assignments. We report the mean difference in bias score and the $95\%$ confidence interval. Bolded values indicate bias is amplifying for black hair color and italic values indicate bias is amplifying for blonde hair.}
        \vskip 0.15in
    \begin{tabular}{lp{0.28\textwidth}p{0.27\textwidth}p{0.27\textwidth}}
    \toprule
         &  1 & 2 & 3\\ 
    \midrule
    \textsc{Original} 
    & \textit{\{\texttt{heavy makeup},\texttt{no beard},\texttt{oval face},\texttt{pale skin},\texttt{pointy nose},\texttt{receding hairline},\texttt{smiling},\texttt{wearing lipstick},\texttt{young}\}}
    & \{\texttt{bags under eyes},\texttt{high cheekbones},\texttt{male},\texttt{mouth slightly open},\texttt{narrow eyes},\texttt{receding hairline},\texttt{smiling},\texttt{wavy hair}\} 
    &  \{\texttt{bags under eyes},\texttt{bangs},\texttt{eyeglasses},\texttt{male},\texttt{narrow eyes},\texttt{wavy hair}\}\\
    & \\
    & $100.0 \pm 0.0$ & $100.0 \pm 0.0$ & $100.0 \pm 0.0$\\ 
    & \\
    & \\
    \textsc{Oversample} 
    & \{\texttt{bags under eyes},\texttt{high cheekbones},\texttt{male},\texttt{mouth slightly open},\texttt{narrow eyes},\texttt{receding hairline},\texttt{smiling},\texttt{wavy hair}\} 
    & \textit{\{\texttt{arched eyebrows},\texttt{heavy makeup},\texttt{high cheekbones},\texttt{mouth slightly open},\texttt{no beard},\texttt{oval face},\texttt{receding hairline},\texttt{smiling},\texttt{straight hair},\texttt{wearing lipstick},\texttt{wearing necklace},\texttt{young}\}} 
    & \textit{\{\texttt{heavy makeup},\texttt{no beard},\texttt{oval face},\texttt{pale skin},\texttt{pointy nose},\texttt{receding hairline},\texttt{smiling},\texttt{wearing lipstick},\texttt{young}\}}\\
    & \\
    & $100.0 \pm 0.0$ & $100.0 \pm 0.0$ & $100.0 \pm 0.0$\\
    & \\
    & \\
    \textsc{RBA} 
    & \textit{\{\texttt{bangs},\texttt{big nose},\texttt{eyeglasses},\texttt{high cheekbones},\texttt{mouth slightly open},\texttt{no beard},\texttt{pointy nose},\texttt{smiling},\texttt{wearing lipstick},\texttt{wearing necklace}\}} 
    & \textit{\{\texttt{bangs},\texttt{big nose},\texttt{eyeglasses},\texttt{high cheekbones},\texttt{mouth slightly open},\texttt{no beard},\texttt{pointy nose},\texttt{smiling},\texttt{wearing lipstick}\}}
    & \{\texttt{bags under eyes},\texttt{big nose},\texttt{high cheekbones},\texttt{male},\texttt{mouth slightly open},\texttt{no beard},\texttt{receding hairline},\texttt{smiling},\texttt{wavy hair}\}\\
    & \\
    & $100.0 \pm 0.0$ & $100.0 \pm 0.0$ & $90.0 \pm 17.5$\\
    & \\
    & \\
    \textsc{Adv} 
    & \textbf{\{\texttt{arched eyebrows},\texttt{bags under eyes},\texttt{bangs},\texttt{big nose},\texttt{heavy makeup},\texttt{high cheekbones},\texttt{mouth slightly open},\texttt{no beard},\texttt{smiling},\texttt{straight hair},\texttt{wearing earrings},\texttt{wearing lipstick}\}} 
    &  \{\texttt{bags under eyes},\texttt{male},\texttt{no beard},\texttt{wavy hair},\texttt{wearing earrings}\}
    & \textbf{\{\texttt{arched eyebrows},\texttt{bangs},\texttt{big nose},\texttt{heavy makeup},\texttt{no beard},\texttt{straight hair},\texttt{wearing earrings},\texttt{wearing lipstick},\texttt{young}\}} \\
    & \\
    & $87.5 \pm 0.0$ & $83.3 \pm 0.0$ & $83.1 \pm 17.5$\\
    & \\
    & \\
    \textsc{DomInd} 
    & \textit{\{\texttt{big nose},\texttt{eyeglasses},\texttt{high cheekbones},\texttt{mouth slightly open},\texttt{no beard},\texttt{receding hairline},\texttt{smiling},\texttt{wearing earrings},\texttt{wearing necklace},\texttt{young}\}}
    & \{\texttt{bangs},\texttt{eyeglasses},\texttt{heavy makeup},\texttt{high cheekbones},\texttt{mouth slightly open},\texttt{no beard},\texttt{smiling},\texttt{wearing lipstick}\}
    & \textit{\{\texttt{bags under eyes},\texttt{big nose},\texttt{eyeglasses},\texttt{male},\texttt{narrow eyes},\texttt{no beard},\texttt{smiling}\}}\\
    & \\
    & $100.0 \pm 0.0$ & $68.6 \pm 5.0$ & $64.8 \pm 30$ \\
    & \\
    & \\
    \textsc{Data Repair} 
    & \textbf{\{\texttt{high cheekbones}, \texttt{male}, \texttt{mouth slightly open}, \texttt{oval face}, \texttt{receding hairline}, \texttt{wearing earrings}, \texttt{young}\}}
    & \{\texttt{arched eyebrows}, \texttt{big nose}, \texttt{heavy makeup}, \texttt{high cheekbones}, \texttt{no beard}, \texttt{rosy cheeks}, \texttt{smiling}, \texttt{straight hair}, \texttt{wavy hair}, \texttt{wearing lipstick}, \texttt{young}\} 
    & \textit{\{\texttt{arched eyebrows}, \texttt{heavy makeup}, \texttt{no beard}, \texttt{oval face}, \texttt{pale skin}, \texttt{straight hair}, \texttt{wavy hair}, \texttt{wearing lipstick}, \texttt{young}\}}
    \\
    & \\
    & $100.0 \pm 0.0$ & $100.0 \pm 0.0$ & $100.0 \pm 0.0$\\
    \bottomrule
    \end{tabular}
    \label{tab:group_qual_celeba}
        \vskip -0.1in
\end{table*}

\section{Qualitative results}
\label{sec:app_qual}
We consider the top attributes or groups of attributes that contribute to bias amplification. Concretely, we examine the difference in bias scores between predictions and the training set when calculating $\text{BiasAmp}_\text{MALS}$ and $\text{Multi}_{\text{MALS}}$. The results on the ``balanced'' datasets from Sec.~\ref{sec:balance} and mitigation methods from Sec.~\ref{sec:mitigate} are found in Tbls.~\ref{tab:balanced_qual}, \ref{tab:group_qual}, and \ref{tab:group_qual_celeba}.

\smallsec{``Balanced'' models} We find that multi-attributes surfaces a different set of attributes than when only considering for single attributes. In Tbl.~\ref{tab:balanced_qual}, there is some overlap between single attributes in the multi-attribute groups (e.g., \texttt{bus}, \texttt{dog}, \texttt{receding hairline}) for COCO and CelebA. However, for imSitu, the multi and single attributes are disjoint.

\smallsec{Bias mitigation methods} Next, we look at the groups surfaced after applying mitigation techniques (Tbl.~\ref{tab:group_qual},~\ref{tab:group_qual_celeba}). We note certain groups of attributes occur across many mitigation methods. For example, in COCO, \{\texttt{car, motorcycle, truck, handbag}\} occurs in \textsc{RBA}, \textsc{Adv}, and \textsc{DomInd}. Similarly, in imSitu, \{\texttt{indoors, crying}\} also occurs in \textsc{Oversample} and \textsc{RBA} as well as \textsc{Original}. This indicates there are certain groups of attributes which may be more difficult to debias than others. Learning to address the bias amplification arising from these more difficult groups can be a fruitful direction for future work in this space.

\section{Runtime analysis} 
Naively implemented, the runtime for calculating the multi-attribute metric could be exponential as we could iterate through all possible combinations of attributes $\mathcal{A}$ of size $\lvert m \rvert$ for $\lvert m \rvert \in \{1, \dots |\mathcal{A}|\}$. This is especially costly given that $|\mathcal{A}|$ can be large for many visual datasets. However, we only consider the groups of multiple attributes that exist in the dataset; many groups of multiple attributes do not occur in either the training or the test set. Rather, our implementation can run in $\mathcal{O}(n)$ time where $n$ is equal to the number of instances in either the training or test set.

\end{document}